%% file: main.tex
\documentclass[runningheads]{llncs}
\usepackage[T1]{fontenc}
\usepackage{amsmath,amssymb}
\usepackage{booktabs}
\usepackage{longtable}
\usepackage{xcolor}
\usepackage{soul}
\usepackage{graphicx}
\usepackage{microtype}

\usepackage{subcaption}
\usepackage{chemformula}
\usepackage{xstring}
\usepackage{seqsplit}
\usepackage{wrapfig}
\usepackage[hidelinks]{hyperref}
\usepackage{tikz}
\usepackage[colorinlistoftodos,textwidth=2.2cm]{todonotes}
\soulregister\cite7
\soulregister\citep7
\soulregister\citet7
\soulregister\ref7

\makeatletter
\renewcommand\section{\@startsection{section}{1}{\z@}{-10\p@ \@plus -2\p@ \@minus -2\p@}{6\p@ \@plus 2\p@ \@minus 2\p@}{\normalfont\large\bfseries\boldmath\rightskip=\z@ \@plus 8em\pretolerance=10000 }}
\renewcommand\subsection{\@startsection{subsection}{2}{\z@}{-8\p@ \@plus -2\p@ \@minus -2\p@}{4\p@ \@plus 2\p@ \@minus 2\p@}{\normalfont\normalsize\bfseries\boldmath\rightskip=\z@ \@plus 8em\pretolerance=10000 }}
\makeatother

\begin{document}

\title{Agentic Optimization: LLM Agents Select and Configure Algorithms for Engineering Design}

\titlerunning{Agentic Optimization}
%
\author{Dominic Mashak\inst{1,2}\orcidID{0009-0009-2477-1778} \and
Jacob Schrum\inst{1}\orcidID{0000-0002-7315-0515}}
\authorrunning{D. Mashak and J. Schrum}
\institute{Southwestern University, Georgetown TX 78626, USA\\
\email{\{mashakd,schrum2\}@southwestern.edu}
\and
Texas A\&M University, College Station TX 77843, USA\\
\email{mashak@tamu.edu}}
\maketitle

\begin{abstract}
Engineers solving new multi-objective design problems commonly commit to a multi-objective evolutionary algorithm (MOEA) before seeing a 
result. The best 
MOEA differs from benchmark to benchmark. We propose \emph{agentic optimization}, placing an LLM agent outside the optimization loop, where it reads a natural-language problem description to configure and launch algorithms, drawing on parametric 
knowledge rather than on a performance history that a novel problem lacks. We compare five LLMs in a fixed harness on four engineering benchmarks and a real-world SARS-CoV-2 molecular docking task, using hypervolume (HV) and IGD\textsuperscript{+} with multiple comparison-corrected tests. Three models (GPT, DeepSeek, Qwen) dominate every baseline on combined HV and IGD\textsuperscript{+}, and each is significantly better than the best baseline on at least one benchmark, sometimes by applying the same baseline algorithm they surpass, showing the importance of both algorithm configuration and selection. Agent behavior is strongly model-dependent, but even the weakest agents dominate three of the four baselines
overall.
\end{abstract}

\section{Introduction}
\label{sec:intro}

Engineering design problems are often multi-objective and constrained: a car body must be light and safe, a power grid cheap to run and stable, a drug candidate potent and synthesizable.
Multi-objective evolutionary algorithms (MOEAs) are often applied to
such problems~\cite{coello2007book}, but users must commit to one before seeing a
result, and many do so by convention. NSGA-II~\cite{deb:tec2002} is a standard choice across engineering applications~\cite{kiymaz:ichora2025,guan:ecm2021,chouraqui:2015,gulec:ame2023,babaei:est2016,suleyman:smo2025,chaturvedi:buildings2020,cai:ast2022}, largely because it is familiar and
well tested. However, by the No Free Lunch Theorems, no optimizer is best on every problem~\cite{wolpert:ieeetec1997}, and different MOEAs win on different benchmarks~\cite{zitzler:ec2000,khare:lncs2003,zapotecas:eaai2023}. Our own baselines confirm this: under the same budget, population size and 
constraint handling,
the best of four MOEAs changes from problem to problem.
This calls for per-problem algorithm selection. The usual formulation is data-driven: a selector is trained on instance features and 
a performance history gathered by running 
algorithms
over many instances of a 
problem family~\cite{kerschke:ec2019}, and recent work replaces hand-designed features with representations learned by large language models (LLMs)~\cite{wu:ijcai2024}. An engineer with a novel problem 
has no such history, and on expensive
problems, building the history would consume the entire budget. We instead ask whether the parametric knowledge an LLM acquired in training can make that choice before the first evaluation is spent. Choosing the algorithm is only half of the decision. Extending automated algorithm configuration to combined algorithm selection and hyperparameter optimization (CASH)~\cite{thornton:kdd2013}, where algorithm choice becomes part of the search space, has been identified as a promising direction for expensive engineering optimization~\cite{long:smo2025}.


We propose \emph{agentic optimization}, which places the model \emph{outside} the optimization loop.
Most work coupling LLMs with evolutionary computation puts the model inside it, generating metaheuristics~\cite{stein:tec2025} or steering the operators of a fixed algorithm, as in the agentic-guided A-NSGA-II~\cite{zhang:cie2026} and GA-Agent~\cite{narimani:arxiv2026}; in both, the agent acts within a single run of one algorithm. Our agent never writes an algorithm or proposes a candidate solution. It
reads a natural-language description of the problem and emits
campaign configurations, each naming an established MOEA, population size, evaluation
budget and constraint handling,
and it reads each campaign's outcome before deciding whether to continue or switch. Because the agent draws on the same algorithm pool
and budget as the baselines it is compared against, any difference in outcome
comes from its configurations.

We apply algorithm selection on a per-problem basis.
Agentic algorithm selection should be expected to perform as well as the best baseline algorithm simply by making the correct choice at the outset, and although this is a reasonable ceiling to aim for, the agents sometimes surpass it, either by selecting a different algorithm and configuration for each of several campaigns and carrying the solutions found by one campaign into the next, or by simply choosing a better configuration of the best baseline algorithm.


We
compare five LLMs in a fixed harness against every MOEA in the
pool, on four engineering benchmarks and a novel SARS-CoV-2 molecular docking task, using hypervolume (HV) and IGD\textsuperscript{+} with
multiple-comparison-corrected tests. Three models (GPT, Qwen and DeepSeek) dominate every baseline on combined HV and IGD\textsuperscript{+}, and each is significantly better than the best baseline
on at least one benchmark. Some of these wins come from agents running the same algorithm as the baseline they beat, 
but with a better configuration.
Agent behavior depends strongly on the model, but every agent dominates at least three of the four baselines overall, without the engineer needing to know in advance which algorithm suits the problem. We release the harness as an open-source library, so that practitioners can apply agentic optimization to their own problems (link withheld for review).



\section{Related Work}
\label{sec:related_work}

 
Algorithm selection is commonly framed as a data-driven problem \cite{kerschke:ec2019}: portfolio methods model algorithmic performance from observed runs
\cite{silverthorn:ut2012}, and landscape analysis supplies instance features for trained selectors \cite{beham:thesis2019}. Both require many runs over instances of the same family, which is affordable when evaluations are cheap but not on expensive optimization problems. Wu et al.\ replace hand-designed features by having a pretrained LLM encode the source code of each candidate algorithm, improving on selectors that use problem features alone \cite{wu:ijcai2024}, and follow-up work extends this to feature-free selection \cite{wu:mind2025}. These approaches remain supervised, with a selector fitted against measured performance; we remove the fitting step by asking the model to decide directly.
 
 
Sinha et al.~\cite{sinha:arxiv2026} is the closest existing work to ours. They switch among four acquisition functions for constrained multi-objective Bayesian optimization on synthetic benchmarks and two materials design case studies, using either a UCB multi-armed bandit or a team of LLM agents as the controller. Their
controller chooses which acquisition function proposes each evaluation, and their LLM agents
see only a numeric summary of campaign
progress. Our agent instead configures entire MOEA campaigns, choosing the algorithm, population size, evaluation budget, and constraint handling,
from a natural-language description of the problem, so its first decision draws on the model's prior knowledge before any evaluation is spent. Like their controllers, our agent adapts online thereafter, reading the results of each campaign before configuring the next.
 
A recent survey \cite{liu:arxiv2024} identifies four roles for LLMs in optimization: optimizer, predictor, extractor, and
designer, and treats algorithm selection as a separate, high-level management
task, which is the role our agent occupies. In the designer role, LLaMEA has an LLM write and iteratively refine complete metaheuristics against benchmark feedback \cite{stein:tec2025}. Such an approach may find algorithms superior to any existing one, but at a high evaluation cost, whereas selecting among known algorithms, as we do, leverages what the LLM already knows about them.

\section{Methodology}
\label{sec:methods}


\subsection{Agent Architecture}

\begin{figure}[t]
    \centering
    \begin{minipage}[c]{0.48\linewidth}
        \centering
        \includegraphics[width=\linewidth]{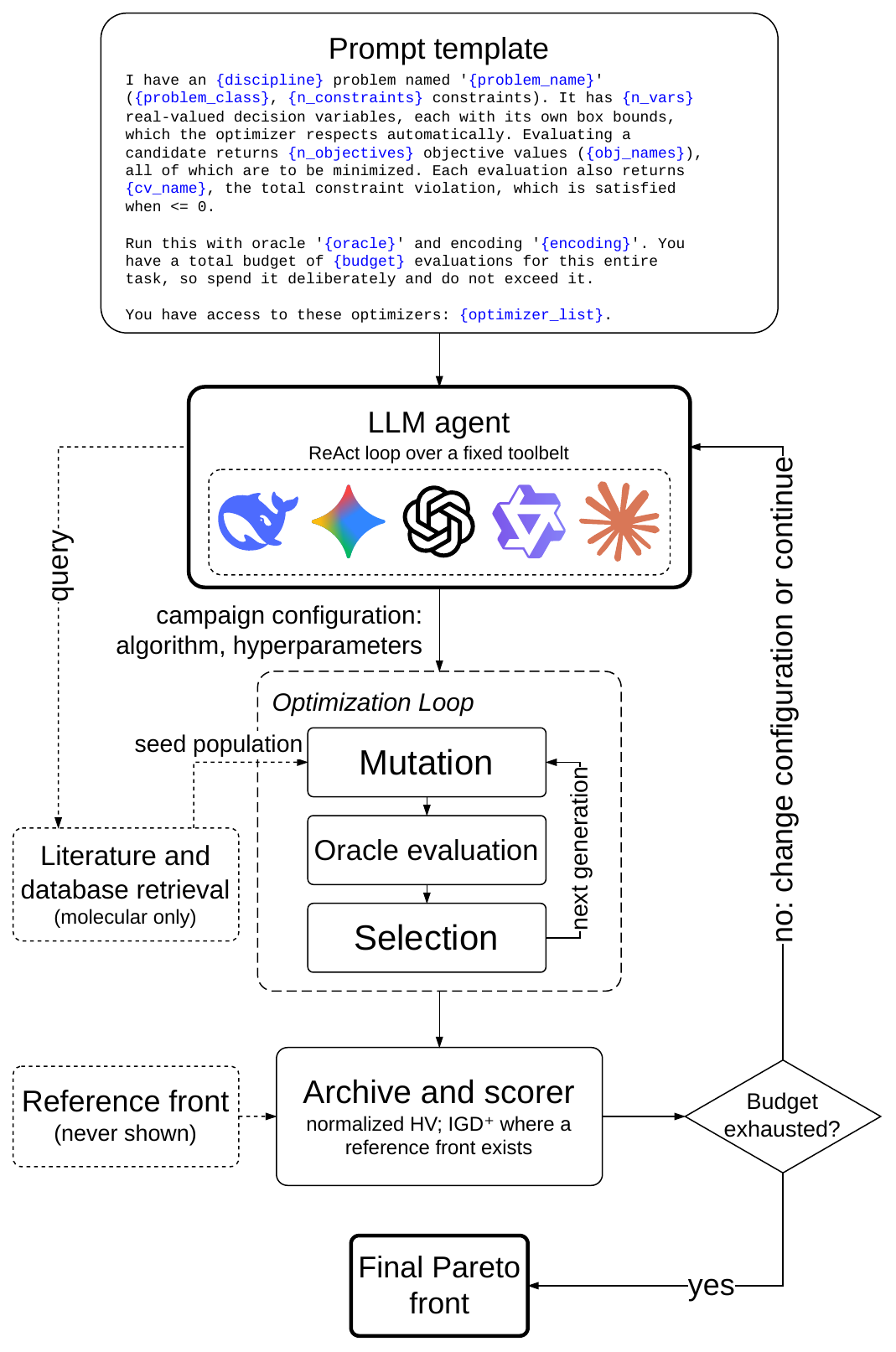}
    \end{minipage}\hfill
    \begin{minipage}[c]{0.51\linewidth}
        \caption{The agentic optimization loop. A problem configuration fills the prompt template, which the LLM agent then receives as input. The agent occupying the outer loop emits a campaign configuration and never enters the optimization loop itself. An inner optimization loop executes the campaign, and everything the agent subsequently sees comes back through the solution archive. The reference front behind IGD\textsuperscript{+} is never revealed to the agent. Literature and database retrieval are only available on the docking benchmark; they provide intelligently selected starting populations for campaigns. The outer loop repeats, continuing the campaign or switching to a different algorithm, until the entire budget is spent.}
        \label{fig:agentic_loop}
    \end{minipage}
\end{figure}

The agent is a ReAct loop \cite{yao:arxiv2023} over a fixed set of tools (Fig.~\ref{fig:agentic_loop}). It receives a natural-language description of the problem, reasons about it, calls tools, reads the results, and stops after the budget is exhausted. Its only ability to influence the outcome is through campaign configurations. 
Each campaign is allotted a share of that budget by the agent. The inner optimization loop runs the chosen MOEA until the allotment is spent or the run's remaining budget is exhausted, scores the resulting solutions, and writes them to the solution archive. Between campaigns, the agent examines the archive and its indicator scores before configuring the next campaign, which may continue with the same algorithm or switch to another.


The agent does not write a configuration as free text. Each affordance is exposed as a tool with a typed input schema 
which names the algorithm, population size, objectives, constraints, and budget as fields, with the permitted values enumerated. The model therefore emits a structured tool call whose arguments are validated before anything runs, meaning 
the prompt needs no instructions about configuration format. A call naming an unregistered algorithm or an out-of-range value is rejected and returned to the agent as an error rather than executed.

Agents are capped at 80 ReAct iterations, where a step is one round of model reasoning followed by a tool call. Sampling and reasoning settings are not uniform across providers; we set only what each provider accepts. Qwen runs locally at temperature 0 with thinking disabled. Claude runs with adaptive extended thinking enabled, which is incompatible with a temperature setting. DeepSeek and Gemini receive neither a temperature nor a reasoning parameter and run at their provider defaults. Qwen is therefore the only agent whose sampling we fix; the hosted models are not deterministic even where a temperature could be set.

\subsection{Campaign Configuration}

Throughout this paper, \emph{oracle} refers to a procedure that evaluates a candidate and returns its objective and constraint values. We use this term rather than ``objective function'' because on some of our benchmarks (Section~\ref{sec:benchmarks}), the procedure is a numerical simulation or external program rather than a closed-form expression. Every oracle is fully deterministic. The optimizer interacts with each oracle in the same way, submitting candidate solutions and receiving numeric results.

Three constraint modes are available. A \texttt{penalty} constraint applies a soft subtraction from fitness, proportional to the relative magnitude of the violation: a candidate violating a constraint by a fraction $r$ of the threshold loses $0.25\,r$, summed over violated constraints~\cite{deb:cmame2000}, so a marginal violation costs little and a gross one costs greatly, while the candidate remains eligible for selection. A \texttt{death} constraint assigns worst fitness, so a violating candidate consumes budget but cannot be selected. A \texttt{hard} constraint is enforced during variation, so violating candidates are never generated. The agent may also emit no constraints at all. The baselines run \texttt{penalty} on the aggregated violation. 
So an agent may do better by configuring constraints differently as well as choosing a different algorithm.

\subsection{Prompt Template}

Every agent uses the prompt template 
shown in Fig.~\ref{fig:agentic_loop}, with slots filled from the same benchmark configuration the baseline MOEAs run from. The slots describe the problem class, the number and type of decision variables, the objectives, 
the constraint information when the problem has constraints, the oracle, the encoding, the evaluation budget, and the list of available MOEA optimizers. Everything else is fixed wording.
The template was designed to contain no steering: no population size, no exploration advice, no preference among the algorithms named in the optimizer list, and no description of what a good result looks like.

\subsection{Benchmark Problems}
\label{sec:benchmarks}

We examine five problems, chosen to span evaluation cost, engineering provenance, and a diverse range of disciplines. Four benchmarks are previously published, and the fifth is 
a novel computationally expensive real-world design task.

Four of the five are \emph{real-valued}: a candidate is a fixed-length vector of continuous decision variables, each confined to its own declared interval $[x^{\mathrm{lo}}_i, x^{\mathrm{hi}}_i]$. Their initial populations are drawn by sampling each variable independently and uniformly from that interval. Genomes are stored in the problem's native units throughout. 
The same generator serves the agent and the baseline MOEAs, so both begin from the same distribution, and each seed draws its own independent sample. Where these problems are constrained, how their constraints are combined is part of the benchmark rather than a choice of ours: each provides its component inequalities already normalized, and pymoo sums them into the single constraint violation measure that it passes to the optimizer, following standard practice for constrained evolutionary optimization~\cite{deb:cmame2000}.

The real-valued engineering benchmarks are established problems from the literature. The \textbf{carside} benchmark is a car side-impact crashworthiness problem \cite{deb:tec2009} from pymoo \cite{blank:ieee2020}; 7 variables, 3 objectives, 10 constraints. 
The compliant-actuator design \textbf{ct1} from the MODAct suite of mechanical design problems \cite{picard:ieeetec2021} from pymoo; 20 variables, 2 objectives, 7 constraints. 
The power flow problem \textbf{opf118} 
on the IEEE 118-bus test system is taken from MATPOWER \cite{zimmerman:tpwrs2011}, solved with pandapower's AC power flow \cite{thurner:tpwrs2018}; 
106 variables, 3 objectives, 2 constraints. 
The rocket injector design problem \textbf{re37} 
is from the RE suite of real-world multi-objective problems \cite{tanabe:asoc2020}; 4 variables, 3 objectives, 0 constraints (prompt omits constraint sentence from Fig.~\ref{fig:agentic_loop}).

The fifth problem is novel and warrants more explanation. We label as \textbf{6W63} the task of designing a small molecule that inhibits the SARS-CoV-2 main protease
(M\textsuperscript{pro}). 
Molecular docking predicts how strongly a candidate binds the enzyme's active site, and 6W63 is the Protein Data Bank
structure of M\textsuperscript{pro} with inhibitor X77, which we prefer to the more commonly used 6LU7 because the latter carries a multi-residue ligand defect \cite{mesecar:pdb2020}.
Candidates are encoded as SELFIES strings \cite{krenn:mlst2020}, so the search space is discrete and of variable
length. There are 3 objectives. The docking score, minimized, is computed by \textit{smina} \cite{koes:jcim2013}, a fork of AutoDock Vina \cite{trott:jcc2010}, using the default Vina scoring function. Drug-likeness (QED, maximized) \cite{bickerton:natchem2012} and synthetic accessibility (SA, minimized) \cite{ertl:jcheminf2009} are computed in RDKit \cite{landrum:rdkit2010}. There are 4 inequality constraints: molecular weight between 300 and 550\,Da, a partition coefficient $\log P \le 5$ ($P$ is the ratio of a molecule's concentration in octanol to that in water and so measures lipophilicity, estimated here by atomic contributions \cite{wildman:jcics1999}), and zero structural alerts, matched against catalogs of pan-assay interference compounds \cite{baell:jmc2010} and of reactive or otherwise unsuitable substructures \cite{brenk:cmc2008}. The weight and $\log P$ bounds follow standard drug-likeness guidelines \cite{lipinski:addr2001}.
Initial populations are constructed by applying between 5 and 15 random SELFIES token
mutations to scaffolds drawn from a library of twenty drug-like scaffolds~\cite{mashak:gecco2026}, with the run's seed controlling
both. The baseline configuration declares the 300\,Da weight floor as a \texttt{hard} rather than a \texttt{death} constraint,
so that it is enforced during this construction: unconstrained random SELFIES average around 90\,Da, so a death-mode floor would eliminate the entire initial
population. The agent declares its own constraint set with each campaign, and may assign a different mode to the same bound or omit
it.

\subsection{Reference Fronts}
\label{sec:fronts}

IGD\textsuperscript{+}~\cite{ishibuchi:emo2015} measures the distance from a reference front to the obtained front, counting only the objectives in which a solution is worse than the reference point, which makes it weakly Pareto compliant. Because it can worsen during a run, we report each run's best (lowest) value. It depends on the quality of the reference front~\cite{tanabe:tec2020}. 
Two 
benchmarks supply an appropriate reference front: 
ct1 carries a 1113-point front from MODAct and re37 a 1500-point approximated front. 


The opf118 problem and docking 6W63 have no published reference front, and the 65-point reference front provided for carside in \textit{pymoo} was dominated by every agent and baseline MOEA from our testing, 
making it unsuitable for IGD\textsuperscript{+} calculation.
For these problems, we construct a reference set as follows: 
every run is reduced to its own non-dominated front; those fronts are unioned, and the non-dominated subset of that union becomes the reference set. This construction is recommended in the standard treatment of performance assessment for stochastic multi-objective optimizers \cite{knowles:tik2006}, and is used in the COCO platform to build and progressively refine its own reference sets \cite{brockhoff:gecco2016}. Our reference sets are pooled from substantially more runs than reported in our results, including baseline MOEAs and preliminary agent runs.

\subsection{Retrieval}
\label{sec:retrieval}

On the docking benchmark only, we conduct additional experiments where the agent has access to retrieval tools that return machine-readable chemical structures from curated databases. Two modes are used: a structural similarity search against PubChem \cite{kim:nar2023} starting from a given molecule, which returns measured physicochemical properties for similar known compounds; and a ChEMBL \cite{zdrazil:nar2024} target and bioactivity search, which returns known actives for a named protein target together with their measured pChEMBL potency values. Both databases index molecules as SMILES \cite{weininger:jcics1988}, the conventional line notation for chemical structure, which is converted to the SELFIES~\cite{krenn:mlst2020} genome used by the search. 

Retrieved actives enter a campaign through the seed pool rather than through the agent's context (Fig.~\ref{fig:agentic_loop}), warm-starting the population from known binders instead of from scaffolds. The baselines have no retrieval tools, so comparing against them requires giving them comparable starting material. We therefore pooled every active that any agent retrieved in any retrieval-enabled run, and every seeded baseline run draws its initial population from that pool. Therefore, the seed pool is made entirely of molecules the agents found: the agents supplied the starting molecules for the baselines they are compared against. In their own seeded runs, agents issue the retrieval queries themselves and decide which of the returned actives to seed a campaign with. Since the pool holds fewer than the population sizes used in evolution, a seeded run takes all of them and fills the remaining slots using the scaffold-and-mutation construction used in the randomly initialized condition. A seeded population is therefore part known binders and part random scaffolds, and the run's seed controls both which actives are used and the scaffolds that fill the rest.


\subsection{Baselines}

Our baseline MOEAs that the agent selects from are from the \textit{pymoo} library \cite{blank:ieee2020} with default operator settings so that no algorithm benefits from implementation-specific tuning. We employ Non-dominated Sorting Genetic Algorithm II (NSGA-II) \cite{deb:tec2002}, Strength Pareto Evolutionary Algorithm~2 (SPEA2) \cite{zitzler:2001}, S-metric Selection Evolutionary Multi-Objective Algorithm (SMS-EMOA) \cite{hochstrate:ejor2007}, and Adaptive Geometry Estimation-based Multi-Objective Evolutionary Algorithm 2 (AGE-MOEA-II) \cite{panichella:gecco2022}. Importantly, no single one of these algorithms is superior on all benchmark problems; each has its own strengths and weaknesses. This variation in performance motivates the use of an intelligent agent for algorithm selection.

\subsection{Models}

We compare five models with an identical harness: Qwen 3.5 122B \cite{qwen:2026}, run locally at Q4\_K\_M quantization through Ollama 0.33.3 on an AMD Ryzen AI MAX+ 395 with 128\,GB of unified LPDDR5X-8533 memory, of which 96\,GB is addressed to the integrated Radeon 8060S; DeepSeek V4.1 Flash \cite{deepseek:2026}; Claude Sonnet 5.5 \cite{anthropic:2026}; GPT-5.6 Terra \cite{openai:56terra2026,openai:system2026}; and Gemini 3.8 Flash \cite{google:gemini38flash2026}. They span local and API options across most major providers, vary in reasoning behavior, and cover a range of price points, giving a broad view of what is available. Every model receives the same tools, same prompt, and same algorithm pool, so behavioral differences are attributable to the model and nothing else. 

\section{Results}
\label{sec:results}
 

The evaluation budget per run is 50{,}000 on opf118 and 5{,}000 on every other problem. 
Baseline MOEAs use a population of 100 on the real-valued benchmarks and 50 on 6W63, while agents choose their own. We run each baseline 30 times per real-valued benchmark and each agent 10 times, fewer for agents because of API costs. On 6W63, counts vary by agent and condition because several models refused the molecular task as biologically sensitive (see  Fig.~\ref{fig:docking}). Performance is measured by normalized hypervolume (HV)~\cite{zitzler:tec1999} based on a fixed per-objective range for each problem, 
and by IGD\textsuperscript{+} 
(Section~\ref{sec:fronts}). 
For each problem and indicator, a Kruskal-Wallis test 
checks whether any difference exists, and 
if it rejects, 
we apply pairwise Mann-Whitney U tests with Holm-Bonferroni correction.
Effect sizes are given with Vargha-Delaney $\hat{A}_{12}$, the probability that a run of one configuration beats a run of the other, so 0.5 means no difference.
 
\begin{figure}[t]
    \centering
    \includegraphics[width=0.24\linewidth]{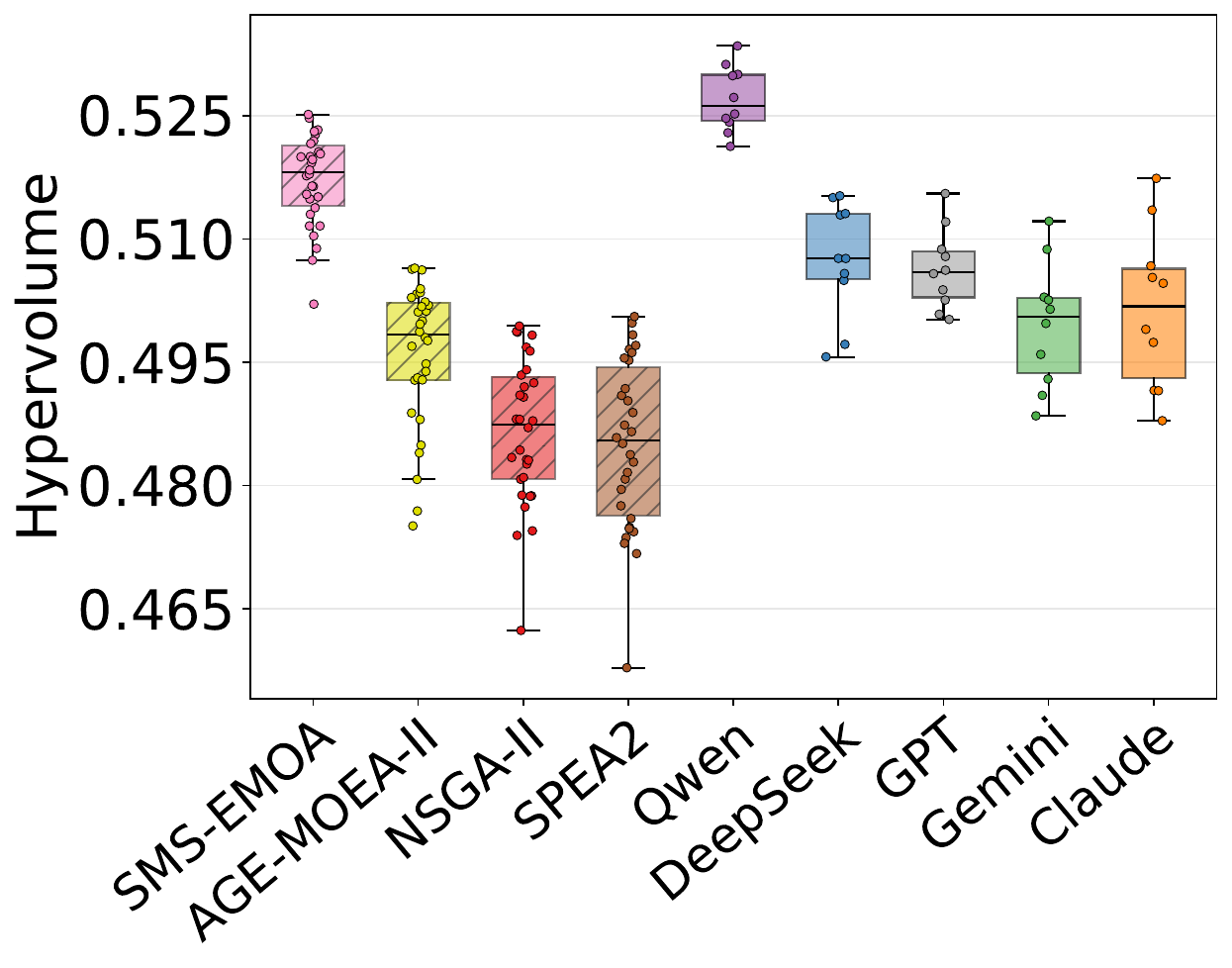}\hfill
    \includegraphics[width=0.24\linewidth]{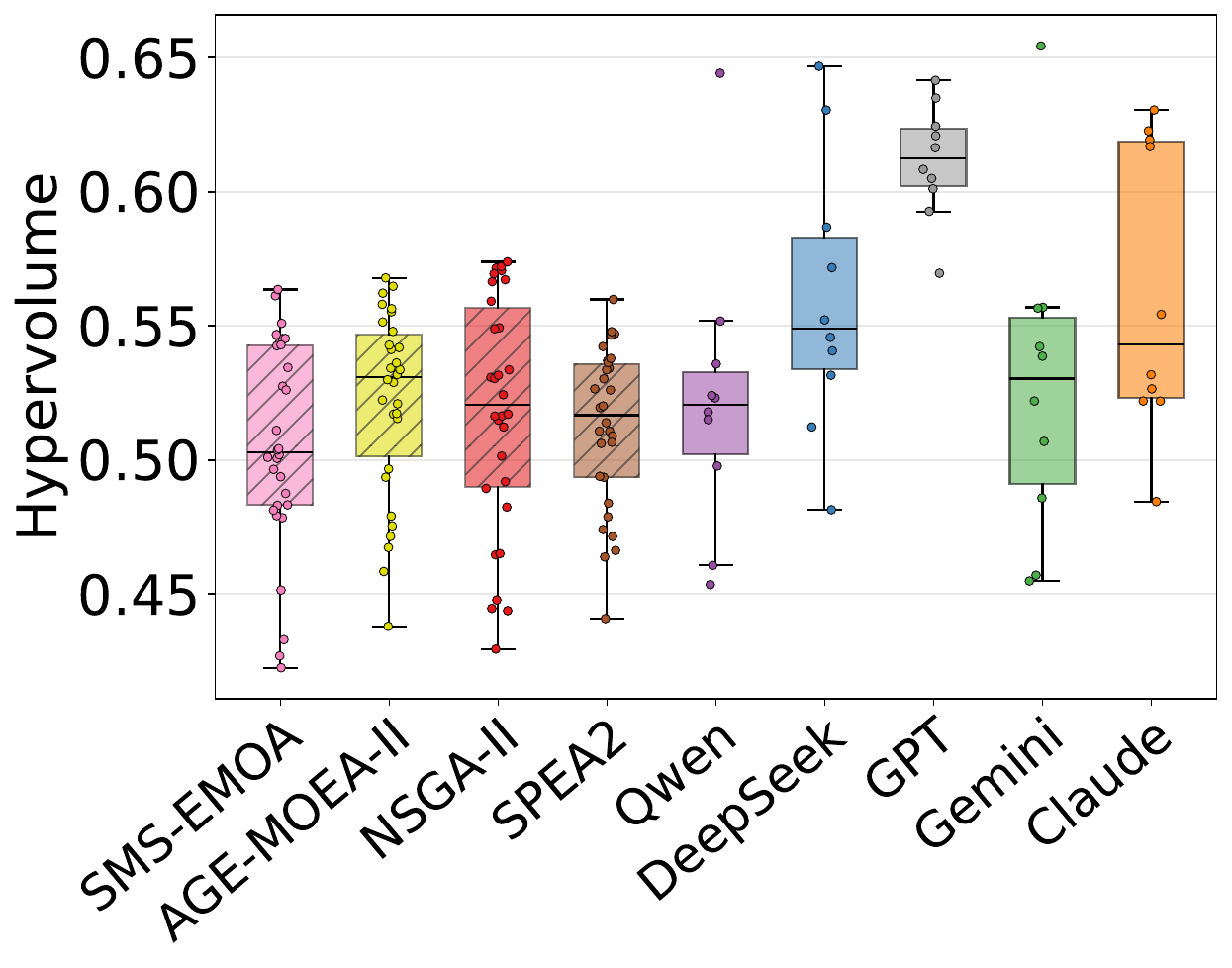}\hfill
    \includegraphics[width=0.24\linewidth]{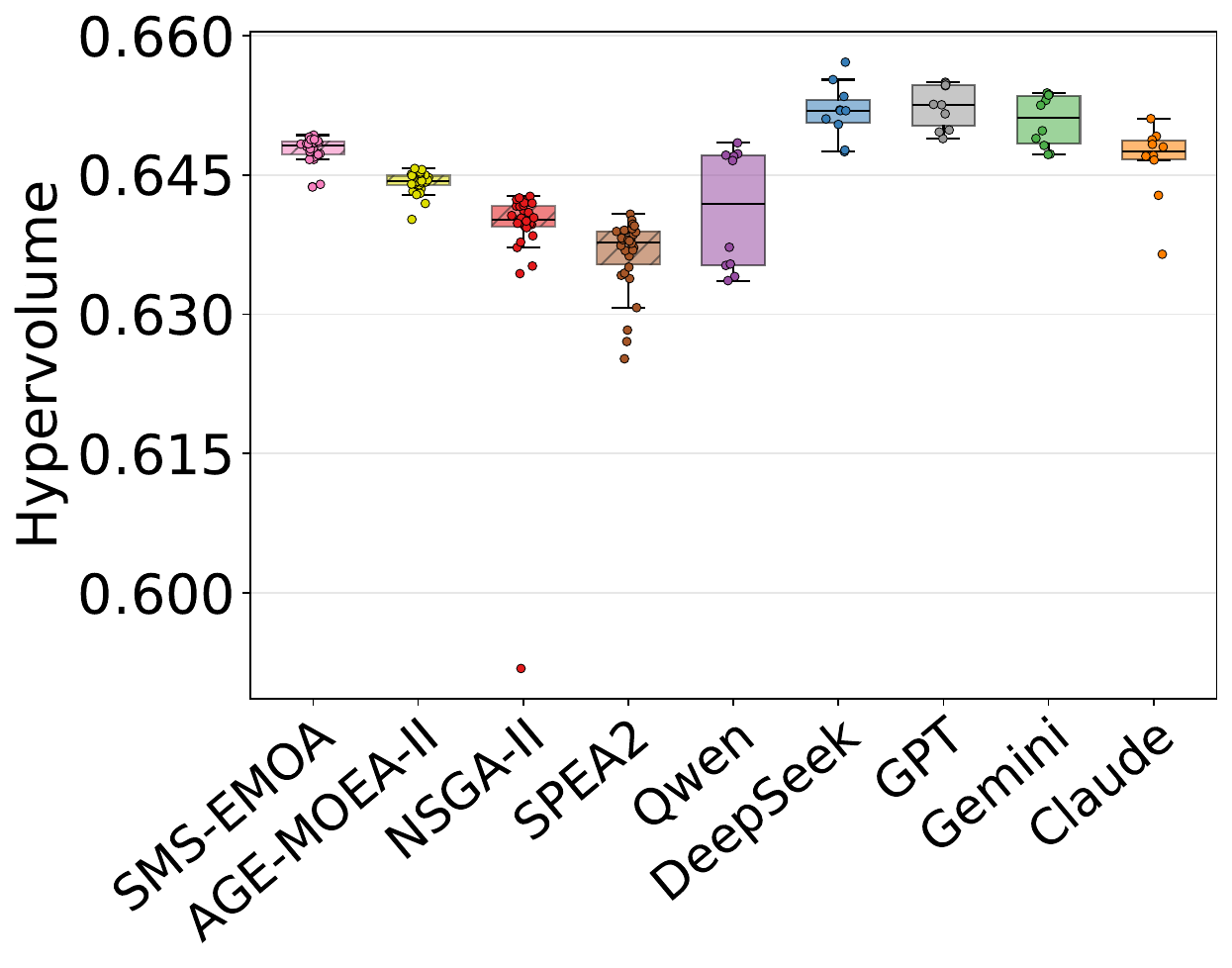}\hfill
    \includegraphics[width=0.24\linewidth]{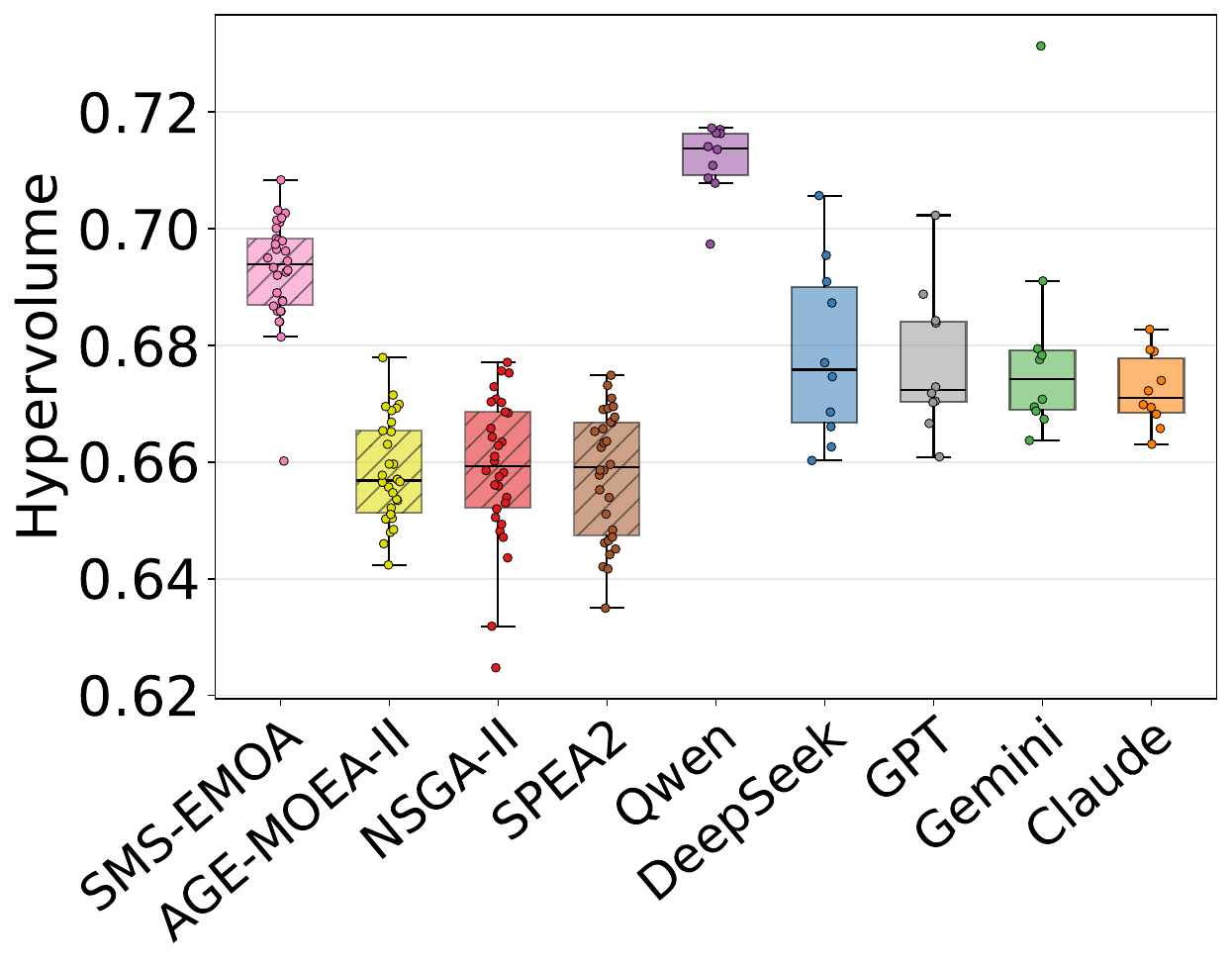}

    \vspace{2pt}
    \begin{subfigure}[t]{0.24\linewidth}
        \centering
        \includegraphics[width=\linewidth]{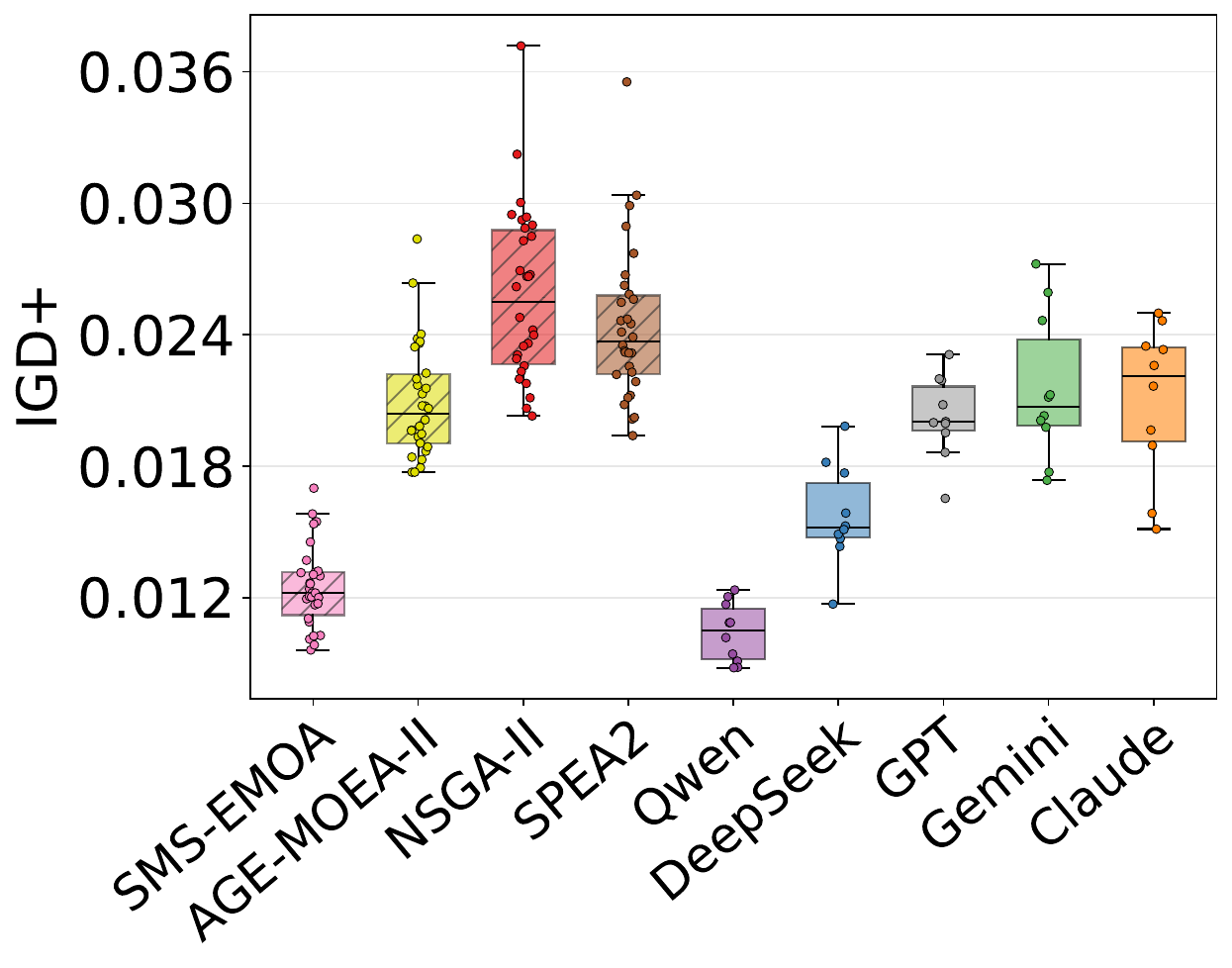}
        \caption{carside}
        \label{fig:carside_box}
    \end{subfigure}\hfill
    \begin{subfigure}[t]{0.24\linewidth}
        \centering
        \includegraphics[width=\linewidth]{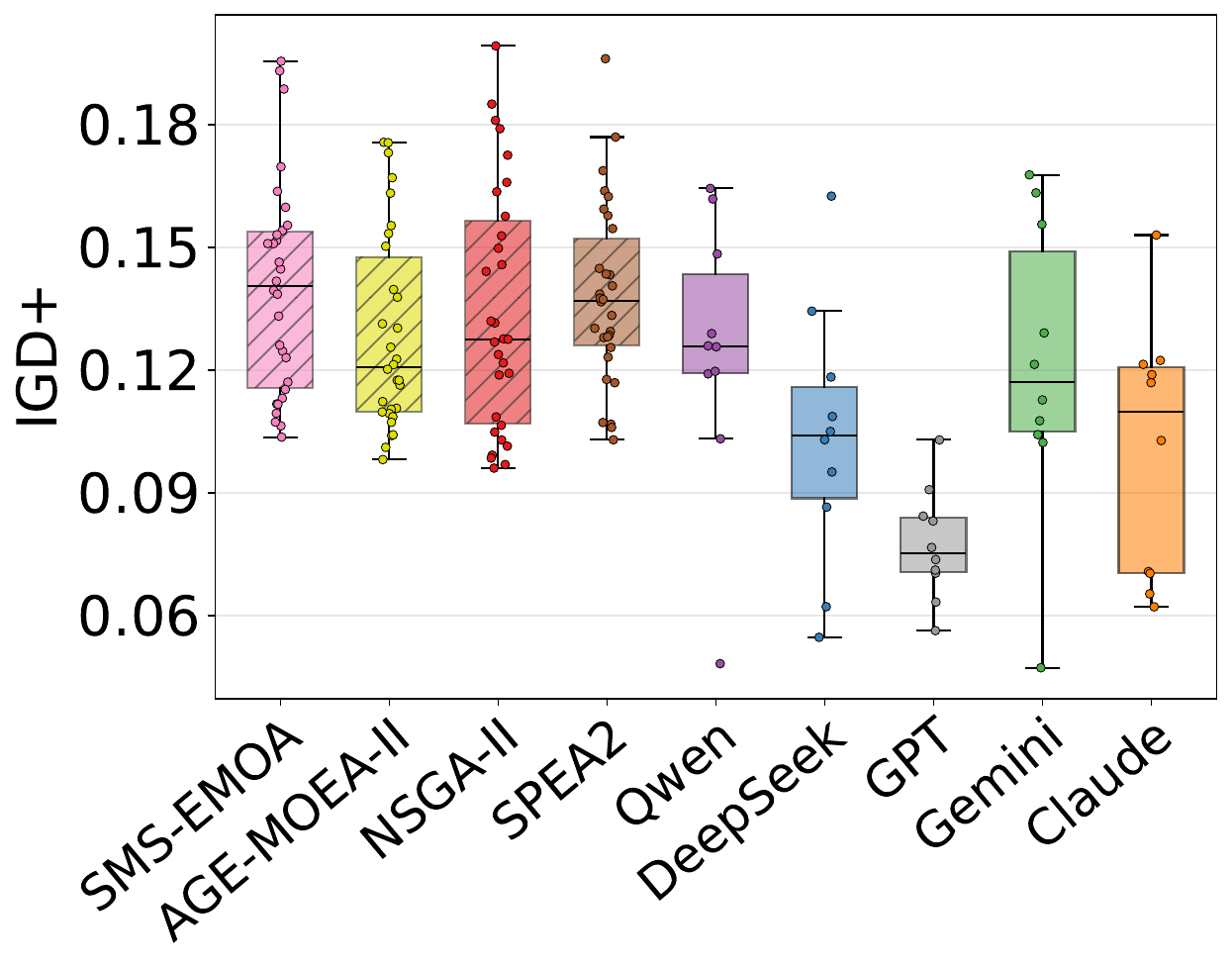}
        \caption{ct1}
        \label{fig:ct1_box}
    \end{subfigure}\hfill
    \begin{subfigure}[t]{0.24\linewidth}
        \centering
        \includegraphics[width=\linewidth]{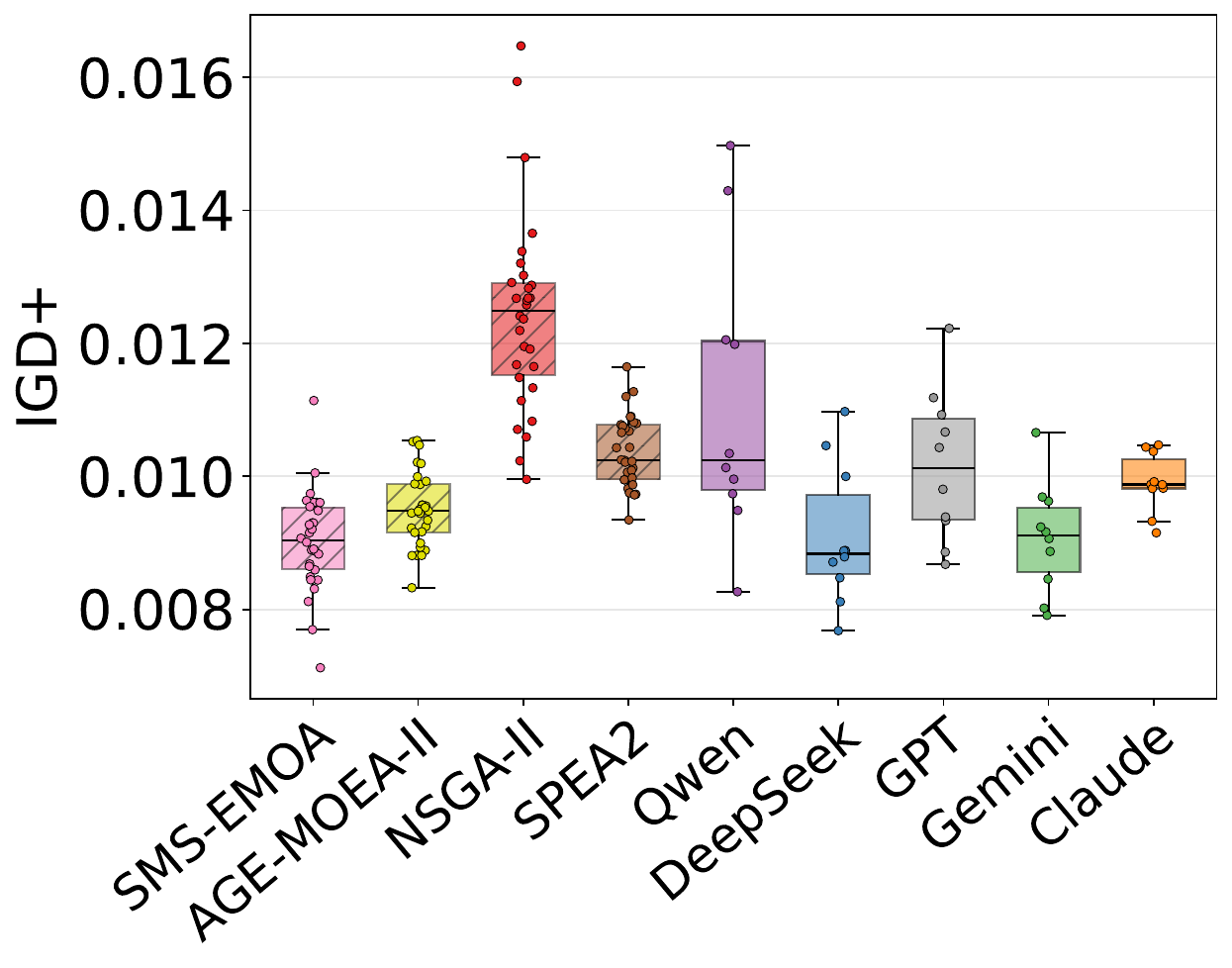}
        \caption{re37}
        \label{fig:re37_box}
    \end{subfigure}\hfill
    \begin{subfigure}[t]{0.24\linewidth}
        \centering
        \includegraphics[width=\linewidth]{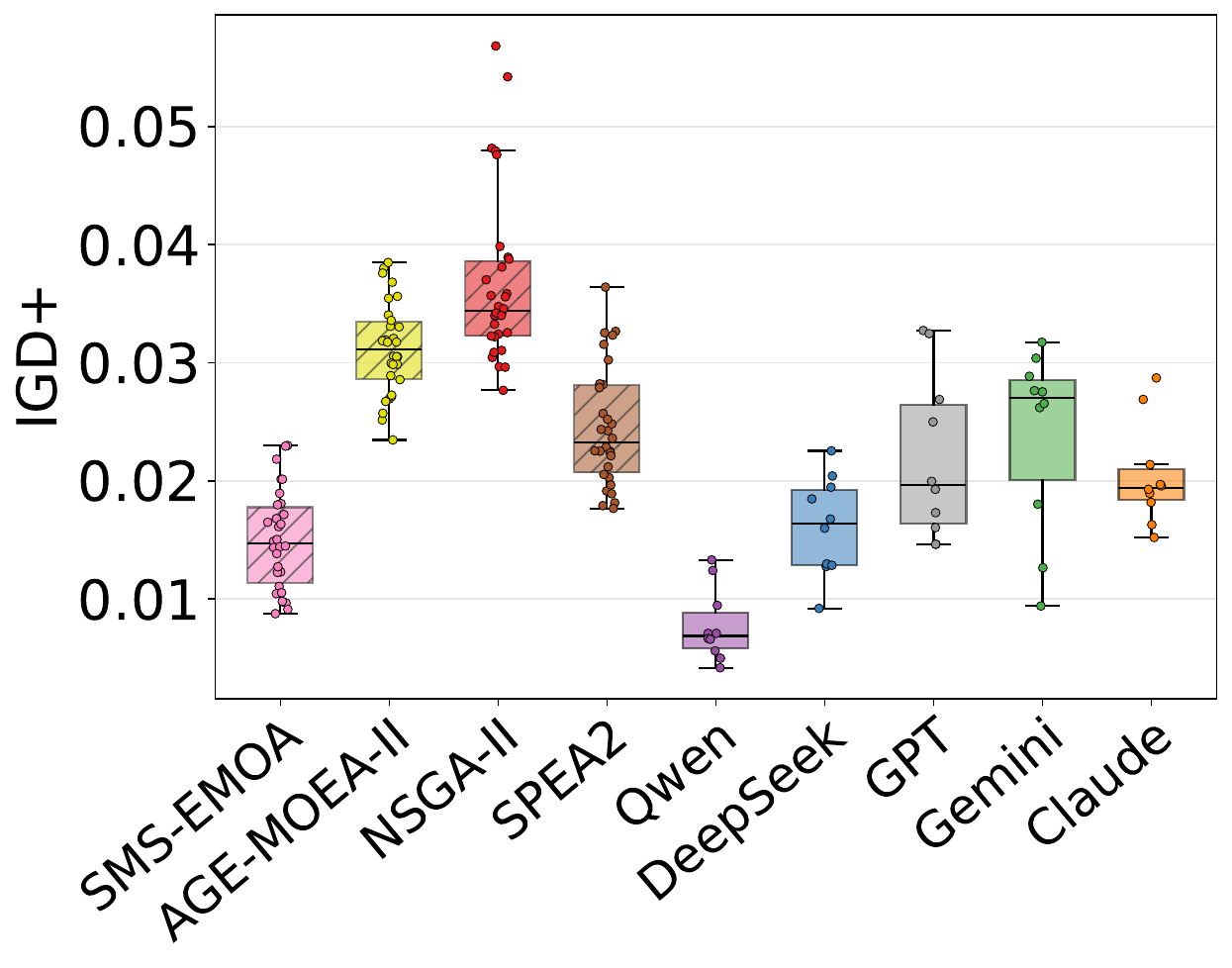}
        \caption{opf118}
        \label{fig:opf118_box}
    \end{subfigure}
    \caption{Final HV (top row, higher is better) and IGD\textsuperscript{+} (bottom row, lower is better; each run's lowest value) on the four real-valued benchmarks. Each point is one run. Hatched boxes are the baseline MOEAs (left; 30 runs each) and solid boxes the agents (right; 10 runs each); whiskers reach 1.5 times the interquartile range. Vertical scales differ between panels.}
    \label{fig:synthetic_box}
\end{figure}
 
\begin{table}[t]
\centering
\caption{Statistical comparison on each problem and indicator (Ind.). $H$ is the Kruskal-Wallis statistic across 
configurations in the row (8 dof 
on real-valued benchmarks, $p<10^{-5}$; 7 and 6 on 6W63, $p<0.01$). Other columns show $\hat{A}_{12}$ of each agent against the best baseline median; values above 0.5 favor the agent. A $^{+}$ or $^{-}$ marks significant differences (after Holm-Bonferroni correction), 
in the agent's favor or against it. A dash marks a model with no runs in that condition.}
\label{tab:a12}
\small
\setlength{\tabcolsep}{4pt}
\begin{tabular}{llrlrrrrr}
\toprule
Task & Ind. & $H$ & Best baseline & Qwen & DeepSeek & GPT & Gemini & Claude \\
\midrule
carside & HV                     & 126.6 & SMS-EMOA    & 0.94$^{+}$ & 0.13$^{-}$ & 0.08$^{-}$ & 0.03$^{-}$ & 0.08$^{-}$ \\
        & IGD\textsuperscript{+}  & 125.3 & SMS-EMOA    & 0.82$^{+}$ & 0.14$^{-}$ & 0.00$^{-}$ & 0.00$^{-}$ & 0.02$^{-}$ \\
ct1     & HV                     & 38.1  & AGE-MOEA-II & 0.43 & 0.71 & 1.00$^{+}$ & 0.50 & 0.67 \\
        & IGD\textsuperscript{+}  & 43.4  & AGE-MOEA-II & 0.48 & 0.76 & 0.99$^{+}$ & 0.55 & 0.72 \\
re37    & HV                     & 137.0 & SMS-EMOA    & 0.14$^{-}$ & 0.88$^{+}$ & 0.99$^{+}$ & 0.79 & 0.42 \\
        & IGD\textsuperscript{+}  & 101.0 & SMS-EMOA    & 0.14$^{-}$ & 0.54 & 0.21 & 0.49 & 0.12$^{-}$ \\
opf118  & HV                     & 110.9 & SMS-EMOA    & 0.96$^{+}$ & 0.24 & 0.16$^{-}$ & 0.16$^{-}$ & 0.04$^{-}$ \\
        & IGD\textsuperscript{+}  & 127.0 & SMS-EMOA    & 0.92$^{+}$ & 0.41 & 0.20 & 0.18$^{-}$ & 0.18$^{-}$ \\
\midrule
6W63    & HV                     & 23.0 & SMS-EMOA    & 0.74 & 0.75 & 0.71 & -- & 0.32 \\
(rand.) & IGD\textsuperscript{+}  & 18.7 & NSGA-II     & 0.48 & 0.60 & 0.47 & -- & 0.23 \\
6W63    & HV                     & 52.6 & SMS-EMOA    & 0.88$^{+}$ & 0.68 & -- & -- & 0.73 \\
(lit.)  & IGD\textsuperscript{+}  & 28.5 & SMS-EMOA    & 0.47 & 0.54 & -- & -- & 0.63 \\
\bottomrule
\end{tabular}
\end{table}
 
The Kruskal-Wallis test rejects on every problem and indicator (Table~\ref{tab:a12}), so the configurations differ everywhere, but how they differ varies by benchmark (Fig.~\ref{fig:synthetic_box}). We take the real-valued benchmarks in turn before combining them. 


On carside, SMS-EMOA beats other baselines on both HV and IGD\textsuperscript{+}, with $\hat{A}_{12}\geq0.99$ against the next best baseline, AGE-MOEA-II. However, Qwen, which ran a single SMS-EMOA campaign in all 10 runs (Fig.~\ref{fig:carside_usage}), is significantly better than SMS-EMOA on both indicators (HV $p<0.001$; IGD\textsuperscript{+} $p=0.04$). 
The other agents fall short of SMS-EMOA on both indicators, 
though they have better medians than the other baselines. DeepSeek, Gemini, and Claude each launched 20 to 22 campaigns, and Gemini and Claude spent half of theirs on NSGA-II. On carside, the best outcome came from committing to the right algorithm.
 
Results differ on ct1 and feature high variation. 
No pair of baselines differs significantly on either indicator ($p=1$ for every pair). 
GPT's HV and IGD\textsuperscript{+} scores are significantly better than the best baseline, AGE-MOEA-II ($p<0.001$ on both). 
GPT launched one campaign per run, split evenly between NSGA-II and AGE-MOEA-II, so its advantage is not from choosing one better algorithm. 
The other agents are not significantly different from AGE-MOEA-II, although 9 of their 40 runs also exceed the best baseline run. 
 

On re37, baselines are highly consistent. 
The baseline order matches carside, with SMS-EMOA 
ahead of AGE-MOEA-II and the rest. 
On IGD\textsuperscript{+}, SMS-EMOA 
and AGE-MOEA-II 
are indistinguishable, but NSGA-II is clearly worst. 
GPT is significantly better than SMS-EMOA in HV ($p<0.001$), as is DeepSeek ($p=0.007$). Gemini's HV beats SMS-EMOA, but not significantly so ($p=0.08$). 
None of the three agents significantly differs from SMS-EMOA on IGD\textsuperscript{+}. 
GPT again used only NSGA-II and AGE-MOEA-II, two baselines that SMS-EMOA beats here. Claude matches SMS-EMOA in HV but is significantly worse on IGD\textsuperscript{+}, and Qwen is significantly worse on both. 
 
opf118 shows the widest gap between baselines. SMS-EMOA's median HV 
exceeds the next best of NSGA-II 
($\hat{A}_{12}=0.98$), while NSGA-II, SPEA2 and AGE-MOEA-II are indistinguishable in HV. On IGD\textsuperscript{+}, SMS-EMOA 
leads SPEA2 and the other baselines. 
As on carside, Qwen ran a single SMS-EMOA campaign in all 10 runs, and is significantly better than the SMS-EMOA baseline on both indicators (HV $p<0.001$, IGD\textsuperscript{+} $p=0.002$).
The other four agents launched 10 (GPT) to 35 campaigns each, mostly on AGE-MOEA-II, NSGA-II and SMS-EMOA. Their median HVs 
are above the three weaker baselines but below SMS-EMOA. 
Gemini nonetheless produced the single best run on the problem, 
so its switching strategy can work but does not do so reliably.
 
\begin{figure}[t]
    \centering
    \begin{subfigure}[t]{0.155\linewidth}
        \centering
        \includegraphics[width=\linewidth]{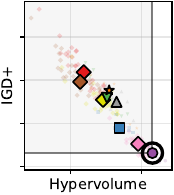}
        \caption{carside}
        \label{fig:carside_tradeoff}
    \end{subfigure}%
    \hfill
    \begin{subfigure}[t]{0.155\linewidth}
        \centering
        \includegraphics[width=\linewidth]{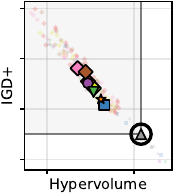}
        \caption{ct1}
        \label{fig:ct1_tradeoff}
    \end{subfigure}%
    \hfill
    \begin{subfigure}[t]{0.155\linewidth}
        \centering
        \includegraphics[width=\linewidth]{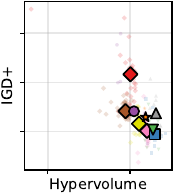}
        \caption{re37}
        \label{fig:re37_tradeoff}
    \end{subfigure}%
    \hfill
    \begin{subfigure}[t]{0.155\linewidth}
        \centering
        \includegraphics[width=\linewidth]{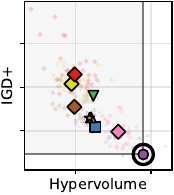}
        \caption{opf118}
        \label{fig:opf118_tradeoff}
    \end{subfigure}%
    \hfill
    \begin{subfigure}[t]{0.155\linewidth}
        \centering
        \includegraphics[width=\linewidth]{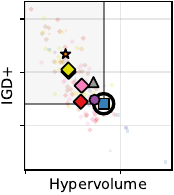}
        \caption{6W63, r. i.}
        \label{fig:docking_rand_tradeoff}
    \end{subfigure}%
    \hfill
    \begin{subfigure}[t]{0.155\linewidth}
        \centering
        \includegraphics[width=\linewidth]{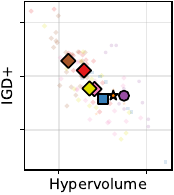}
        \caption{6W63, l. s.}
        \label{fig:docking_lit_tradeoff}
    \end{subfigure}

    \vspace{6pt}
    \begin{minipage}[c]{0.12\linewidth}
        \centering
        \includegraphics[width=\linewidth]{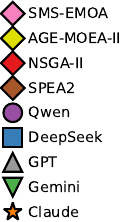}
    \end{minipage}%
    \hspace{4pt}
    \begin{minipage}[c]{0.84\linewidth}
        \raggedright
        \caption{Median HV (higher better) vs.\ median IGD\textsuperscript{+} (lower better) on four real-valued benchmarks (\subref{fig:carside_tradeoff}--\subref{fig:opf118_tradeoff}) and 6W63 (\subref{fig:docking_rand_tradeoff}, \subref{fig:docking_lit_tradeoff}); best configurations lie toward the lower right. Diamonds are baseline MOEAs and other markers are agents (see key); faded markers are individual runs. The configuration best on both medians is circled, and the shaded region shows what it dominates. In (\subref{fig:re37_tradeoff}) and (\subref{fig:docking_lit_tradeoff}) no configuration is best on both, so none is circled. 6W63 run counts are as in Fig.~\ref{fig:docking}.}
        \label{fig:tradeoffs}
    \end{minipage}
\end{figure}

\begin{figure}[t]
    \centering
    \begin{subfigure}[t]{0.155\linewidth}
        \centering
        \includegraphics[width=\linewidth]{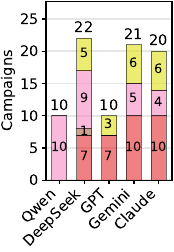}
        \caption{carside}
        \label{fig:carside_usage}
    \end{subfigure}%
    \hfill
    \begin{subfigure}[t]{0.155\linewidth}
        \centering
        \includegraphics[width=\linewidth]{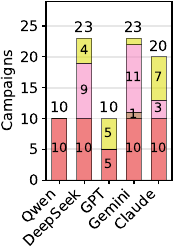}
        \caption{ct1}
        \label{fig:ct1_usage}
    \end{subfigure}%
    \hfill
    \begin{subfigure}[t]{0.155\linewidth}
        \centering
        \includegraphics[width=\linewidth]{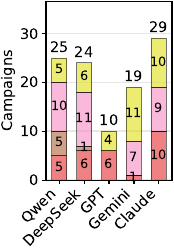}
        \caption{re37}
        \label{fig:re37_usage}
    \end{subfigure}%
    \hfill
    \begin{subfigure}[t]{0.155\linewidth}
        \centering
        \includegraphics[width=\linewidth]{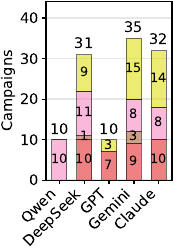}
        \caption{opf118}
        \label{fig:opf118_usage}
    \end{subfigure}%
    \hfill
    \begin{subfigure}[t]{0.155\linewidth}
        \centering
        \includegraphics[width=\linewidth]{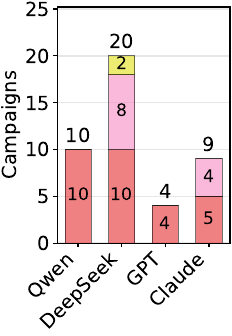}
        \caption{6W63, r. i.}
        \label{fig:docking_rand_usage}
    \end{subfigure}%
    \hfill
    \begin{subfigure}[t]{0.155\linewidth}
        \centering
        \includegraphics[width=\linewidth]{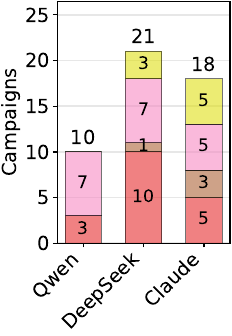}
        \caption{6W63, l. s}
        \label{fig:docking_lit_usage}
    \end{subfigure}

    \caption{Algorithm usage by each agent on real-valued benchmarks (\subref{fig:carside_usage}--\subref{fig:opf118_usage}) and docking in both conditions (\subref{fig:docking_rand_usage}, \subref{fig:docking_lit_usage}). Bars show the total campaigns launched across all runs, stacked by MOEA (colors match Fig.~\ref{fig:tradeoffs}; with all present, bottom-to-top order is NSGA-II, SPEA2, SMS-EMOA, AGE-MOEA-II). Each agent has 10 runs on every real-valued benchmark; 10 means the agent launched one campaign per run and never switched; larger totals mean it split its budget across campaigns. On 6W63, the run counts differ by agent and condition (Fig.~\ref{fig:docking}).}
    \label{fig:usage}
\end{figure}

The trade-offs between median HV and median IGD\textsuperscript{+} (Fig.~\ref{fig:tradeoffs}) show that both are strongly correlated, so configurations fall along a diagonal on every benchmark. The best agent's medians dominate every other configuration's on carside (Qwen), ct1 (GPT) and opf118 (Qwen). On re37, GPT has the highest median HV and DeepSeek the lowest median IGD\textsuperscript{+}, so neither dominates. 
 
Across benchmarks, the agents fall into two behavioral groups (Fig.~\ref{fig:usage}). The first commits to one choice. GPT launches exactly one campaign per run on every benchmark and never chooses SMS-EMOA or SPEA2, splitting its runs between NSGA-II and AGE-MOEA-II. Qwen also commits on three of the four benchmarks, each time to a single algorithm across all 10 runs: SMS-EMOA on carside and opf118, the best baseline on both, and NSGA-II on ct1. The second group explores. DeepSeek, Gemini, and Claude launch between 19 and 35 campaigns per benchmark and use three or four algorithms on each, with DeepSeek having the most even mix. Claude launches NSGA-II in all 10 runs on every benchmark, and Gemini does so in 9 or 10 runs on carside, ct1, and opf118, before moving on mostly to SMS-EMOA and AGE-MOEA-II. This mirrors the convention-driven default described in Section~\ref{sec:intro}, even though NSGA-II is never the best baseline on these problems. SPEA2 is rarely chosen, never more than 5 times on one benchmark.
 
Algorithm choice alone therefore does not explain agent performance. GPT outperforms SMS-EMOA on re37 and every baseline on ct1 using only NSGA-II and AGE-MOEA-II, and Qwen, running only SMS-EMOA, outperforms the SMS-EMOA baseline on carside and opf118. With algorithm and budget matched, the difference must come from the rest of the campaign configuration the agent emits, such as population size or constraint handling. Selecting the algorithm still matters, as carside and opf118 show: the agents that spent most of their budget away from SMS-EMOA did not catch the baseline that ran it throughout.
 
\begin{figure}[t]
    \centering
    \begin{minipage}[c]{0.25\linewidth}
        \centering
        \includegraphics[width=\linewidth]{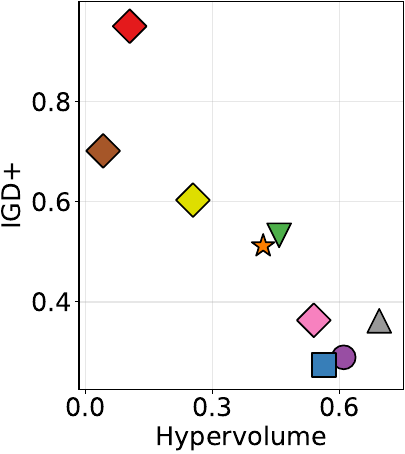}
    \end{minipage}\hfill
    \begin{minipage}[c]{0.73\linewidth}
        \caption{Combined performance over the four real-valued benchmarks. On each benchmark, the nine configurations' median HV and median IGD\textsuperscript{+} are min-max normalized, and the combined score is the mean of the four normalized values: combined HV on the horizontal axis (1 is best) and combined IGD\textsuperscript{+} on the vertical axis (0 is best), so the best lie toward the lower right. The key is in Fig. \ref{fig:tradeoffs}. The non-dominated configurations are GPT, Qwen, and DeepSeek.}
        \label{fig:HV_IGD_combined_front}
    \end{minipage}
\end{figure}

To compare configurations across all four benchmarks at once, we min-max normalize each configuration's median HV and IGD\textsuperscript{+}, 
and take the mean over the four benchmarks (Fig.~\ref{fig:HV_IGD_combined_front}). Normalizing per benchmark gives each benchmark equal weight regardless of its scale. 
Three agents stand out. GPT, Qwen and DeepSeek are the only non-dominated configurations, and each scores at least as well as every baseline on both combined indicators. SMS-EMOA, the strongest baseline, 
is dominated by these three agents, but dominates all other points, including the agents Gemini and Claude, and all other baselines. 
GPT is best on HV followed closely by Qwen, DeepSeek, then SMS-EMOA. 
GPT and SMS-EMOA are tied on IGD\textsuperscript{+}, with Qwen beating both and DeepSeek the best. Although Gemini and Claude are dominated by SMS-EMOA, they still dominate the other baseline MOEAs. 
NSGA-II, the conventional default, has the worst combined IGD\textsuperscript{+} score of any configuration 
and the second-lowest combined HV score. 

 
 
\begin{figure}[t]
    \centering
    \begin{subfigure}[t]{0.24\linewidth}
        \centering
        \includegraphics[width=\linewidth]{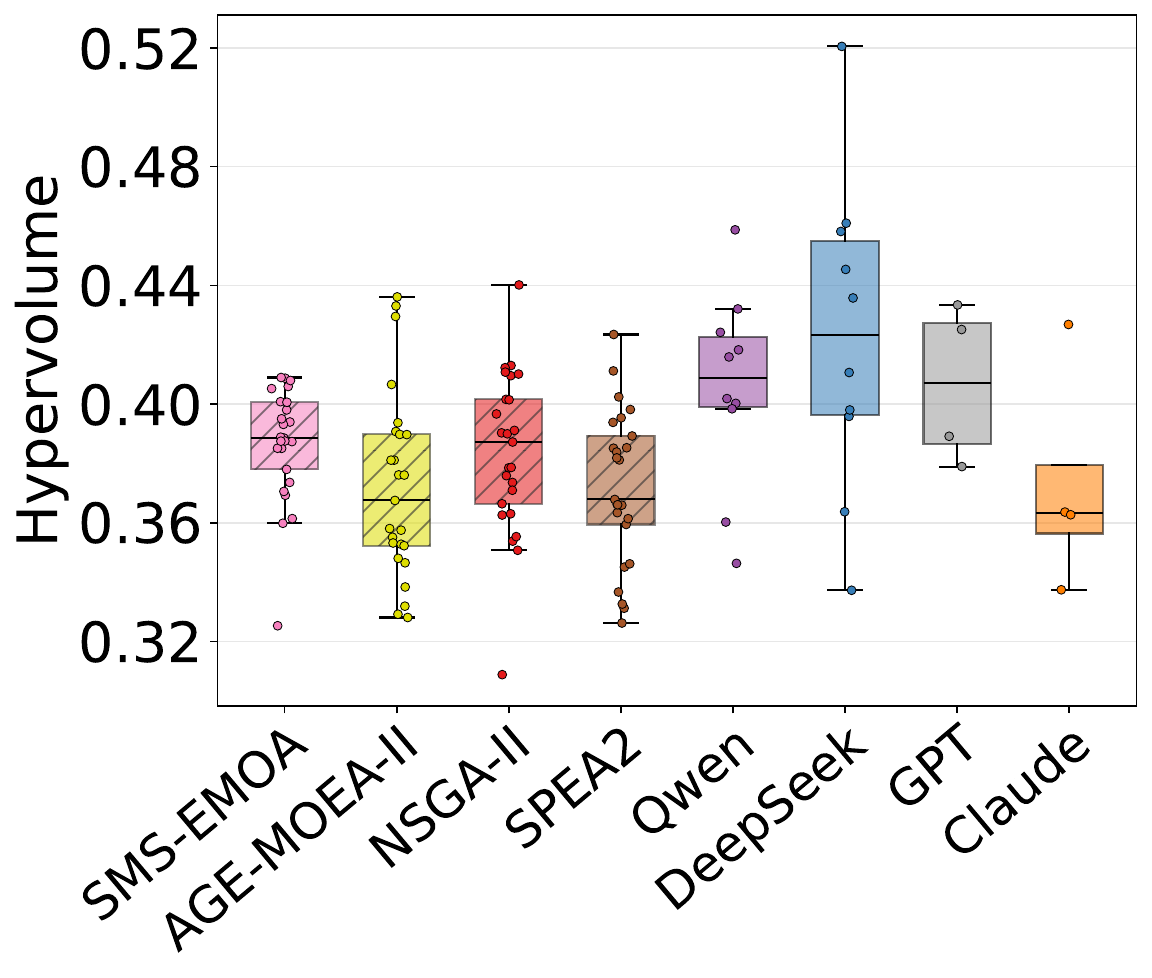}
        \caption{HV, rand. init.}
        \label{fig:docking_rand_hv}
    \end{subfigure}\hfill
    \begin{subfigure}[t]{0.24\linewidth}
        \centering
        \includegraphics[width=\linewidth]{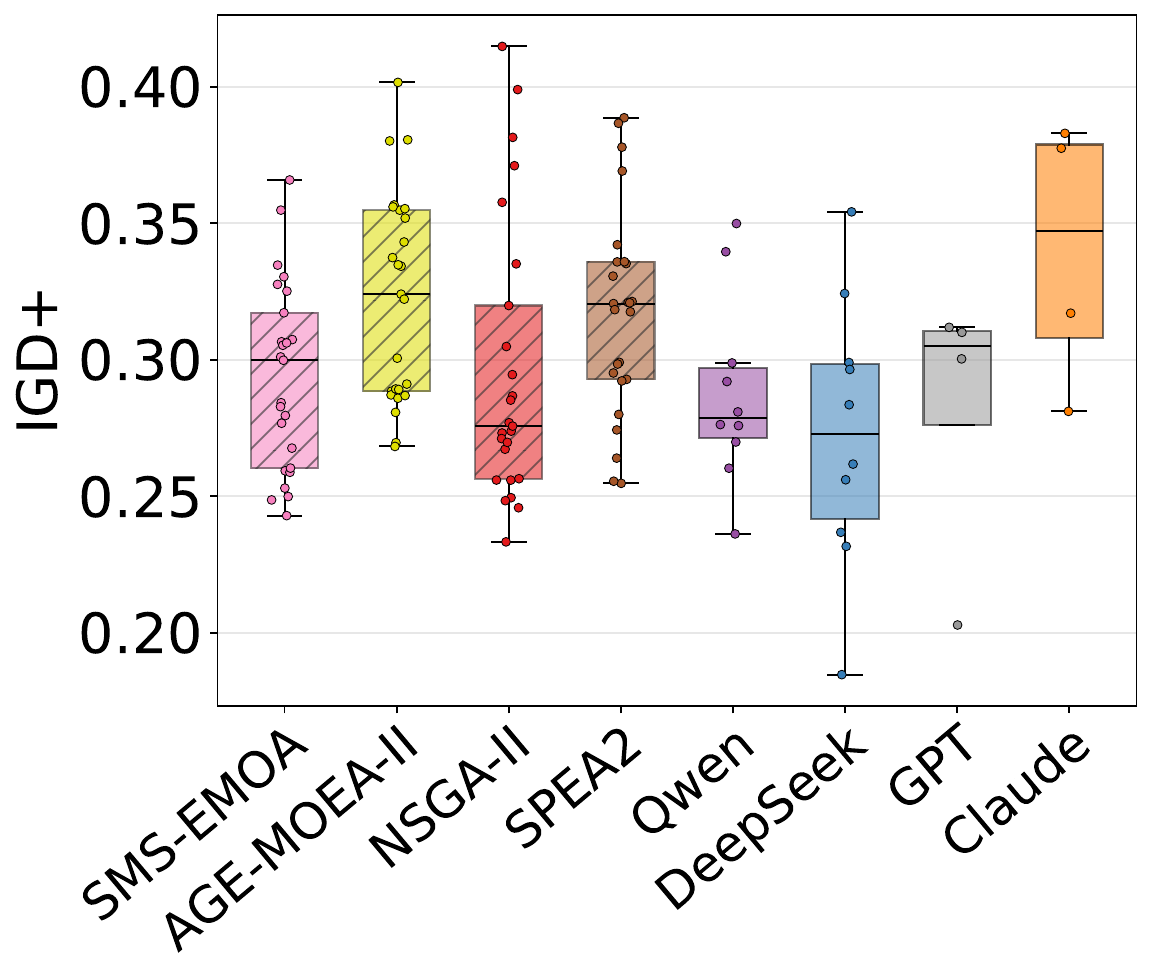}
        \caption{IGD\textsuperscript{+}, rand. init.}
        \label{fig:docking_rand_igd}
    \end{subfigure}\hfill
    \begin{subfigure}[t]{0.24\linewidth}
        \centering
        \includegraphics[width=\linewidth]{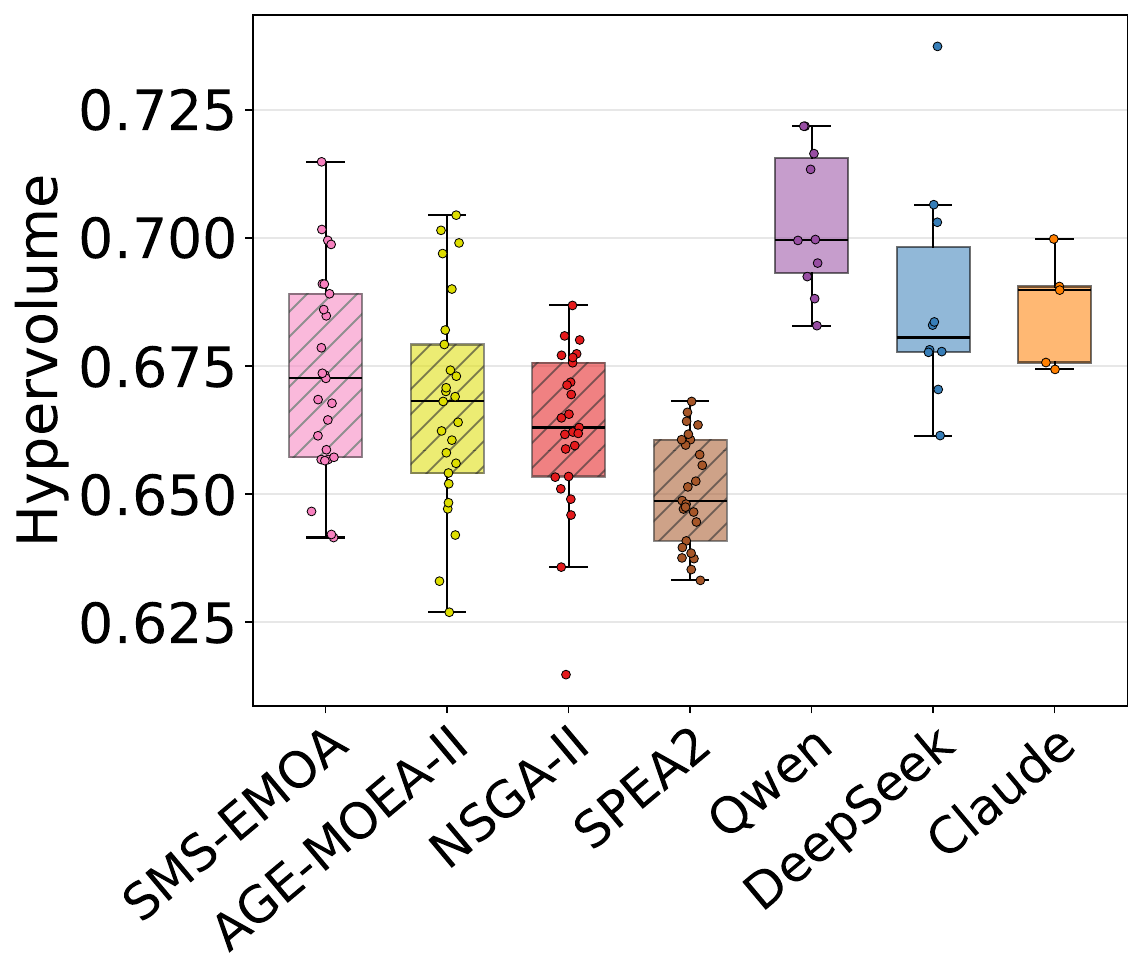}
        \caption{HV, lit. seeds}
        \label{fig:docking_lit_hv}
    \end{subfigure}\hfill
    \begin{subfigure}[t]{0.24\linewidth}
        \centering
        \includegraphics[width=\linewidth]{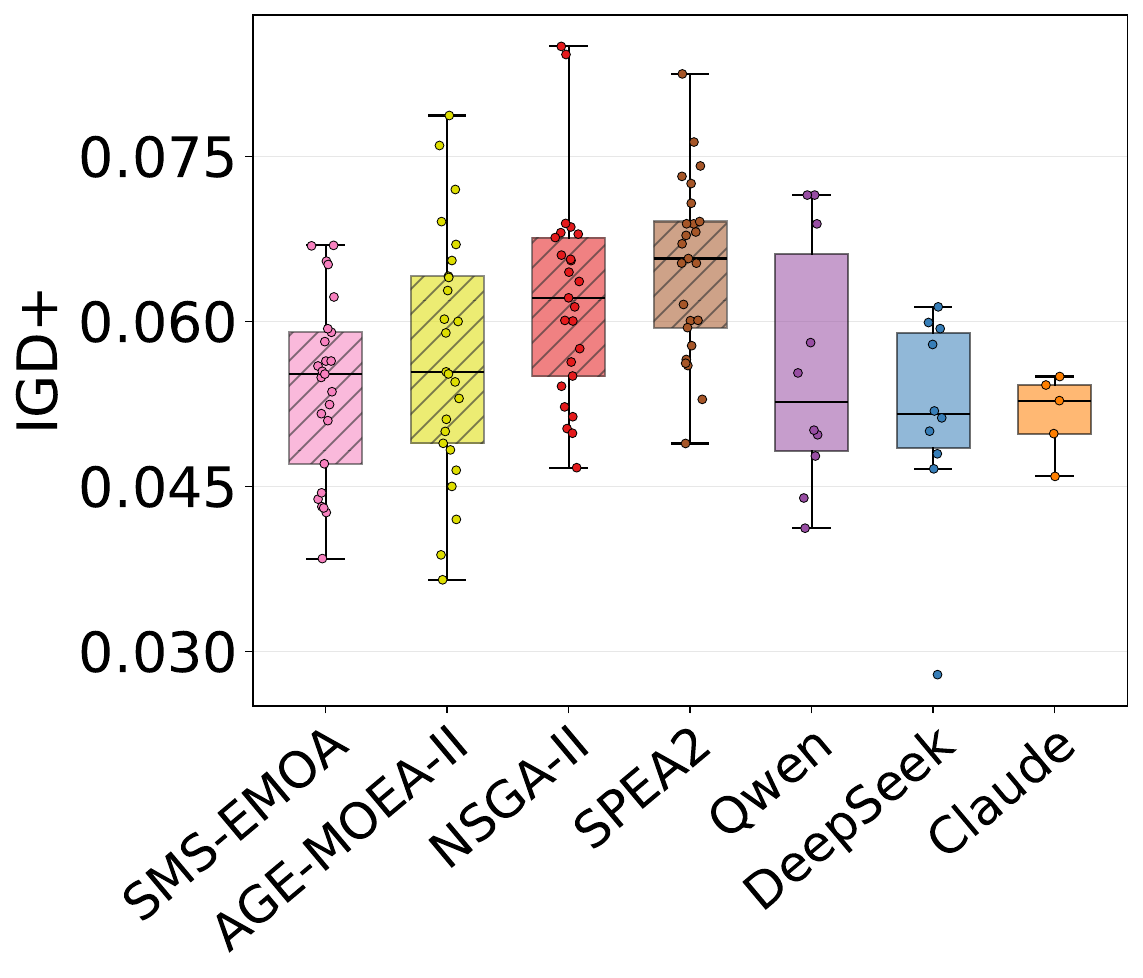}
        \caption{IGD\textsuperscript{+}, lit. seeds}
        \label{fig:docking_lit_igd}
    \end{subfigure}
    \caption{Results on docking against 6W63. (a, b): every run starts from the scaffold-based population (Section~\ref{sec:benchmarks}) with retrieval disabled; baselines have 25 runs each, Qwen and DeepSeek 10, and GPT and Claude 4. (c, d): every run, baselines included, starts from a population seeded with retrieved actives (Section~\ref{sec:retrieval}); baselines have 25 runs each, Qwen and DeepSeek 10, and Claude 5. Missing models and runs are refusals; Gemini refused every run and GPT every seeded run. Both conditions share the same normalization and reference set, so values are comparable across them, but vertical scales differ. Boxes as in Fig.~\ref{fig:synthetic_box}.}
    \label{fig:docking}
\end{figure}
 
6W63 docking differs from the real-valued benchmarks in its discrete SELFIES encoding and its evaluation cost, and we ran it in two conditions (Fig.~\ref{fig:docking}). In the first, every run starts from the scaffold-based population and retrieval is disabled. The Kruskal-Wallis test detects a difference on both indicators (HV: $H=23.0$, $p=0.002$; IGD\textsuperscript{+}: $H=18.7$, $p=0.009$), but no pair of configurations differs significantly after Holm-Bonferroni correction (smallest corrected $p=0.13$, DeepSeek against AGE-MOEA-II in HV). 
Among baselines, SMS-EMOA has the highest median HV, 
while NSGA-II 
has the lowest median IGD\textsuperscript{+}. 
Overall, DeepSeek has the best median on both indicators, 
and produced the best single outlier run. 
Qwen 
and GPT 
also have higher median HV than every baseline, 
but none of the differences are significant under Holm-Bonferroni error correction. 
Claude has the worst medians on both indicators, even losing to baselines. Qwen and GPT launch a single NSGA-II campaign in every run, whereas DeepSeek 
and Claude run a variety of MOEAs (Fig.~\ref{fig:docking_rand_usage}).


Agents and baselines alike start from retrieved actives, seeded as described in Section~\ref{sec:retrieval}. 
Seeding conveys significant advantage, but also causes certain commercial models to refuse requests on the grounds of biological sensitivity. Gemini and GPT refused every seeded request. Claude refused often enough that, despite attempts to reach the intended 10 runs, only 5 completed. The reported seeded runs are therefore every run that completed, not a selection of the better ones. Still, successful seeded runs have higher HV 
and lower IGD\textsuperscript{+} 
than every unseeded run. 
Because the baselines are seeded too, this gap measures the value of the retrieved actives rather than of any agent decision. 
SMS-EMOA is best on both indicators, though nearly tied 
with AGE-MOEA-II. SPEA2 is significantly worst in HV ($p\leq0.04$ against each other baseline). The three 
agents 
have a higher median HV than 
baselines. 
Qwen's HV is significantly better than every baseline, including SMS-EMOA ($p=0.009$), whereas Claude and DeepSeek are significantly better only than NSGA-II and SPEA2. All agents also have lower median IGD\textsuperscript{+} than every baseline, 
although none differs significantly from SMS-EMOA on it ($p=1$ for all three after correction). Claude thus moves from the weakest unseeded configuration to one of the strongest when seeded, while launching 18 campaigns over its 5 runs across all four algorithms, including SPEA2. Qwen, which ran only NSGA-II in the first condition, chose SMS-EMOA in 7 of its 10 seeded runs.

\section{Discussion}
\label{sec:discussion}

Three of the five models dominate every baseline on the combined front, yet an agent that only selects from a fixed pool might be expected to perform at best as well as the best member of that
pool. The usage data rule out the simplest explanation: that the agent picked a better algorithm. GPT beats every baseline on ct1 and SMS-EMOA on re37 while launching only NSGA-II and AGE-MOEA-II, two algorithms that SMS-EMOA itself beats on re37, and Qwen beats the SMS-EMOA baseline on carside and opf118 while running nothing but SMS-EMOA.
Algorithm and budget are matched, so the gain must come from the other fields of the campaign configuration the agent emits. Population size is the most likely
candidate. The baselines run with a population of 100, whereas Qwen sets a population of 50 in the large majority of its campaigns, which doubles the number of generations available within the same
budget, and DeepSeek ranges over far more distinct population sizes than any other model. Part of what the agent contributes is
therefore the hyperparameter tuning a practitioner would otherwise
skip. This does not weaken the practical result, since the agent supplies that tuning without additional evaluations, but it does mean the comparison measures selection and configuration together, and a fair follow-up would compare against baselines
with tuned hyperparameters.
 
The results also qualify the expectation
that agents gain by alternating between algorithms. The two clearest per-benchmark wins come from agents that do not switch: GPT launches exactly one campaign per run on every benchmark, and Qwen commits to a single algorithm on three of the four.
Switching is not harmful in itself, since DeepSeek launches between 19 and 35 campaigns per benchmark and still lies on the combined front, but it splits a fixed budget across
campaigns and widens the spread of outcomes: Gemini produced the single best run on
opf118 but has a median below
SMS-EMOA. For an engineer who will run the optimization once, the median and the spread matter more than the best case, which favors
committing. What separates the weaker agents is where they start. Claude launches NSGA-II in every run on every
benchmark, and Gemini does so in nearly every run on three of the
four, reproducing the convention-driven default described in
Section~\ref{sec:intro}: a model's prior about which MOEA to use is learned from a literature in which NSGA-II dominates engineering
applications. With identical prompts, tools and algorithm pool, the five models range from a single committed campaign per run to more than thirty,
so correct tool use is
necessary but 
not sufficient. This is consistent with Da Ros et al.~\cite{daros:cor2026}, who find that frozen open-weight models encode enough about
an instance to support per-instance algorithm
selection; our results suggest that how well such knowledge reaches the configuration an agent
emits varies considerably between models. No model is uniformly best, with Qwen leading on carside and opf118 and GPT on ct1 and re37, so the choice of model becomes a selection problem in its own right, though a more forgiving one, since three of the five dominate every baseline overall.
 
On docking, the largest effect in the study comes from neither the algorithm nor the model but from the starting population. Seeding with known actives lifts every seeded run above every unseeded one, for baselines and agents alike,
so where prior designs exist, supplying them matters more than the choice of optimizer. The agents still make a meaningful contribution:
all three that ran with seeding have a higher median HV than every seeded baseline,
and Qwen is significantly better than all of them.
 
 
Several limitations bound these conclusions. Each agent is run 10 times per benchmark against 30 runs per baseline, so
non-significant agent differences in Table~\ref{tab:a12}
may reflect limited runs rather than a true tie. The benchmark set is small,
and the conclusions about which agent leads rest on
one or two benchmarks each, so the combined front is best read as evidence that strong agents exist rather than as a ranking of models. The
reference fronts behind IGD\textsuperscript{+} are pooled from
runs that include agent runs (Section~\ref{sec:fronts}), so they may sit closer to the agents' fronts than to the baselines'. API models are
moving targets whose outputs are not fully
reproducible. Finally, several frontier models refuse molecular design tasks frequently or outright due to bio-sensitivity concerns, which limits how far the docking results generalize across providers. Repeated attempts did not yield the intended 10 runs for every model.

\section{Conclusions}
\label{sec:conclusions}

Engineers facing new multi-objective design problems must commit to an MOEA before seeing a result, and our baseline comparison shows that no fixed choice is safe: under matched budgets, the best of four MOEAs changes from benchmark to benchmark. We proposed \emph{agentic optimization}, placing an LLM agent \emph{outside} the optimization loop, where it reads a natural-language problem description and emits campaign configurations for established MOEAs, drawing on its prior knowledge rather than the performance history an engineer with a novel problem lacks. On four real-valued engineering benchmarks, three of the five models (GPT, DeepSeek and Qwen) dominate every baseline on combined HV and IGD\textsuperscript{+}, and each is significantly better than the best baseline on at least one benchmark. These gains are not explained by algorithm choice alone, as GPT and Qwen beat baselines running the same algorithms, suggesting 
hyperparameter configuration is another benefit of agentic optimization.
Committing to a single well-chosen campaign was generally more reliable than switching between algorithms, which divides a fixed budget and widens the spread of outcomes, although not always: DeepSeek, which switched on every problem, lies on the combined front overall
. On molecular docking, seeding the initial population with known actives let all of the MOEAs and agents find far higher-performing molecules, and every agent that ran with seeding had a higher median HV than every seeded baseline. 
Agent behavior was 
model-dependent: the two weaker agents opened almost every run with NSGA-II, reproducing the conventional default, presumably a bias inherited from the engineering literature. The choice of model therefore becomes a selection problem in its own right, but a more forgiving one, since even the weakest agents dominate every baseline except SMS-EMOA. 

Several directions follow from these results. Separating the contributions of selection and configuration requires comparison against baselines with tuned hyperparameters. Our harness already
logs campaigns and reflections to a write-only cross-session memory; making it readable could let agents transfer experience across related problems, a form of meta-learning that suits engineering, where design problems recur in families with shared structure. Retrieval, which supplied the seeds on docking, could likewise supply prior designs to additional challenging real-world problems. A model trained specifically for agentic optimization could also replace the off-the-shelf LLMs used here. Decision models such as Jev~\cite{ling:arxiv2026}, which return typed choices from a predefined set with calibrated probabilities rather than free text, suit this role, since our campaign configurations are already typed tool calls with enumerated values. Finally, adding LLM-generated algorithms from frameworks such as LLaMEA~\cite{stein:tec2025} to the pool would let the agent select, configure, and design the optimizer best suited to a new engineering problem.

\bibliographystyle{splncs04}
\bibliography{references}

\clearpage
\appendix
\input{appendix.tex}

\end{document}

%% file: appendix.tex
\begin{figure}[t]
\centering
    \begin{subfigure}[t]{0.48\linewidth}
        \centering
        \includegraphics[width=\linewidth]{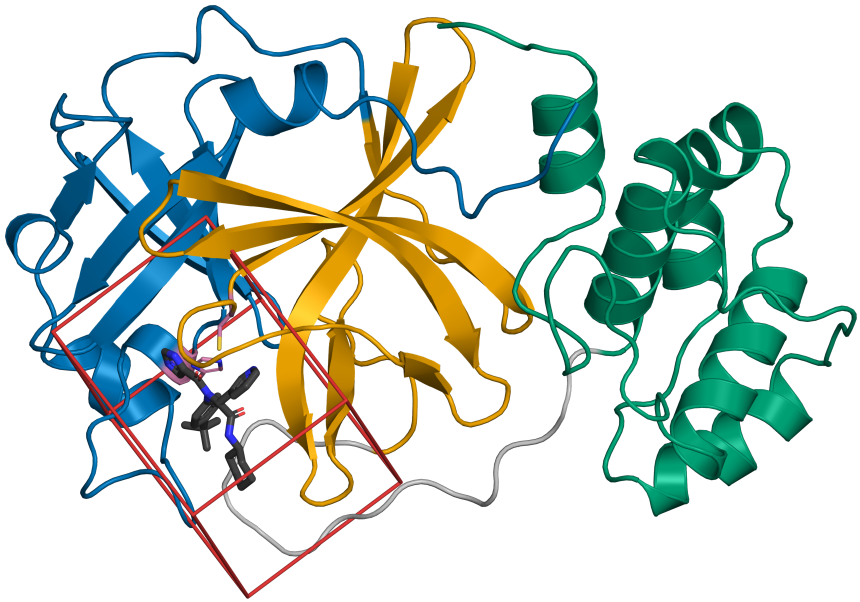}
        \caption{}
        \label{fig:overview}
    \end{subfigure}
    \hfill
    \begin{subfigure}[t]{0.48\linewidth}
        \centering
        \includegraphics[width=\linewidth]{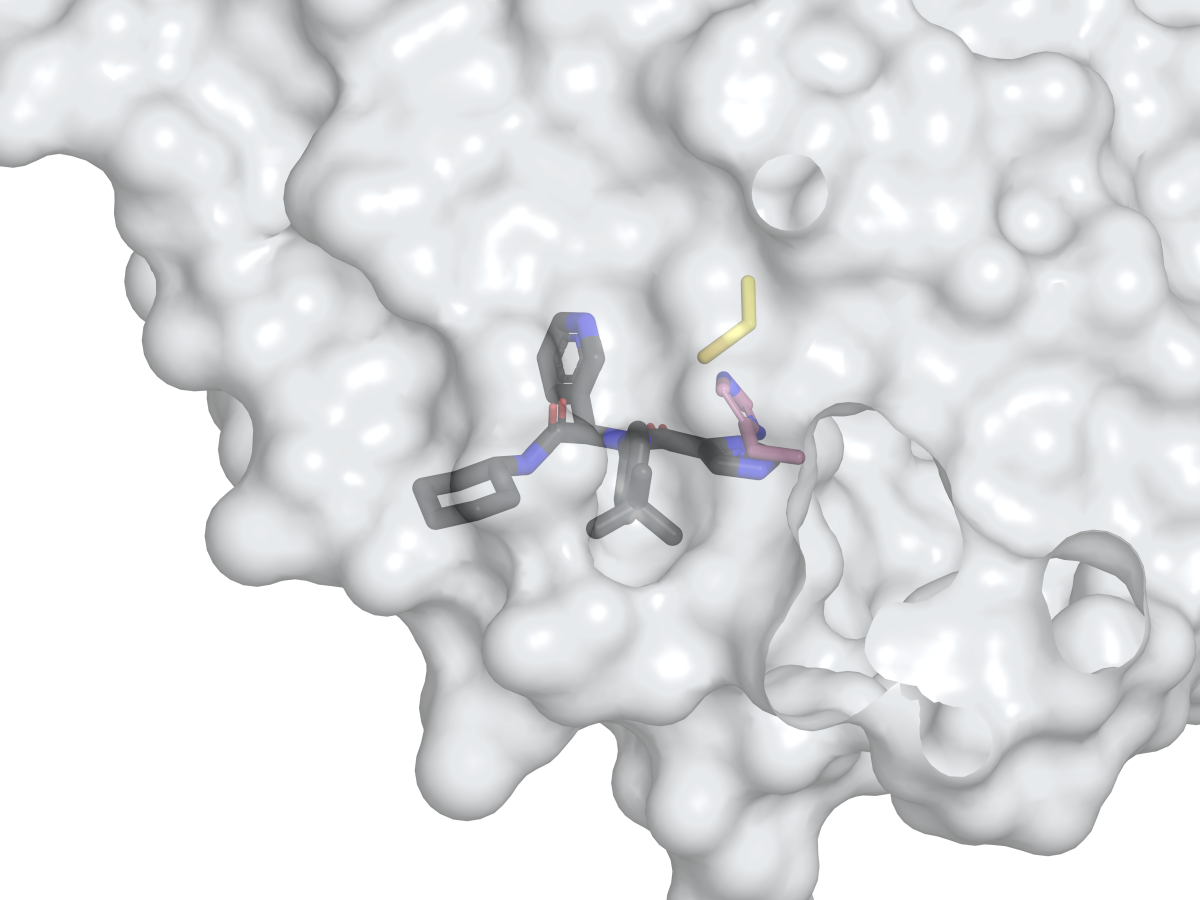}
        \caption{}
        \label{fig:site}
    \end{subfigure}
    \caption{The docking target. \textbf{(a)} The M\textsuperscript{pro} protomer of 6W63 colored by domain: I (blue, residues 1--101), II (orange, 102--184), III (green,
201--306) and the linker (grey), with the co-crystal ligand X77 in dark sticks, the
catalytic dyad His41/Cys145 in purple, and the \textit{smina} search box in red. The box is
recomputed from the receptor files the oracle docks against and checked against the recorded
edge lengths. \textbf{(b)} The same site at 50\,\% surface transparency; X77 is enclosed, so an
opaque surface would hide it. His41 (purple) and Cys145 (yellow) are sticks; the S1 and S2
pocket residues are Met49, Phe140, Asn142, Gly143, His163, Met165, Glu166 and Gln189.}
    \label{fig:6w63_receptor}
\end{figure}

\begin{figure}[!ht]
    \centering
    \begin{subfigure}[t]{0.31\linewidth}
        \centering
        \includegraphics[width=\linewidth]{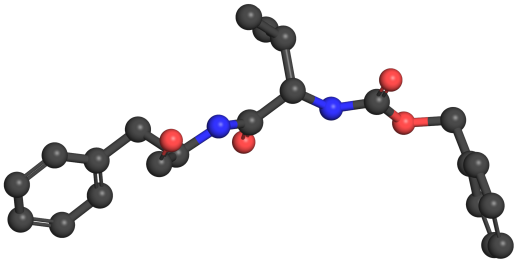}
        \caption{MDL-28170}
        \label{fig:L2}
    \end{subfigure}\hfill
    \begin{subfigure}[t]{0.31\linewidth}
        \centering
        \includegraphics[width=\linewidth]{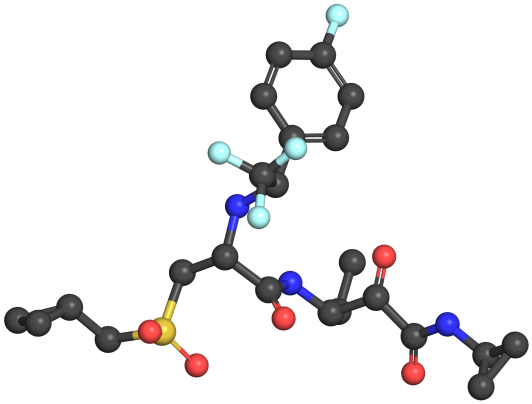}
        \caption{VBY-825}
        \label{fig:L3}
    \end{subfigure}\hfill
    \begin{subfigure}[t]{0.31\linewidth}
        \centering
        \includegraphics[width=\linewidth]{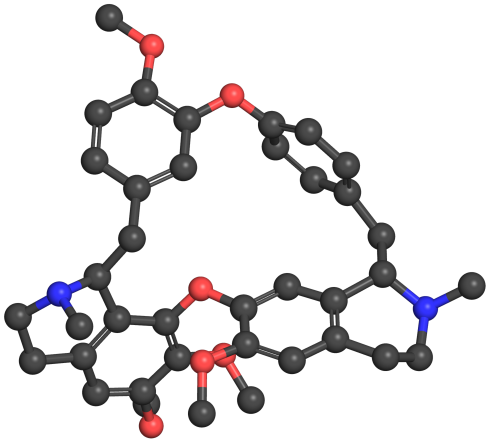}
        \caption{Tetrandrine}
        \label{fig:L4}
    \end{subfigure}\hfill
    \\[4pt]
    \begin{subfigure}[t]{0.31\linewidth}
        \centering
        \includegraphics[width=\linewidth]{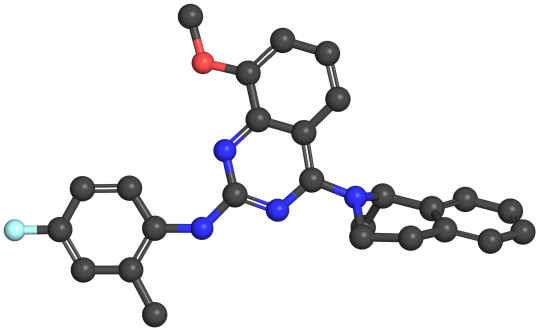}
        \caption{YH-1238}
        \label{fig:L5}
    \end{subfigure}\hfill
    \begin{subfigure}[t]{0.31\linewidth}
        \centering
        \includegraphics[width=\linewidth]{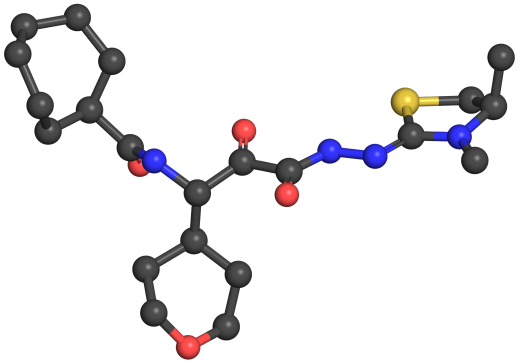}
        \caption{ONO-5334}
        \label{fig:L6}
    \end{subfigure}\hfill
    \begin{subfigure}[t]{0.31\linewidth}
        \centering
        \includegraphics[width=\linewidth]{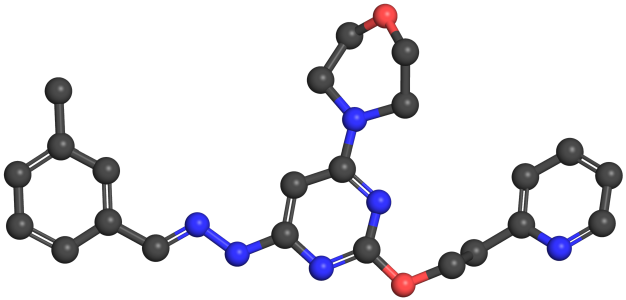}
        \caption{Apilimod}
        \label{fig:L7}
    \end{subfigure}
    \caption{The six best retrieved molecules in our literature seed pool by docking score, QED, and SA score.}
    \label{fig:lit_seeds}
\end{figure}

\begin{figure}[!ht]
    \centering
    \begin{subfigure}[t]{0.31\linewidth}
        \centering
        \includegraphics[width=\linewidth]{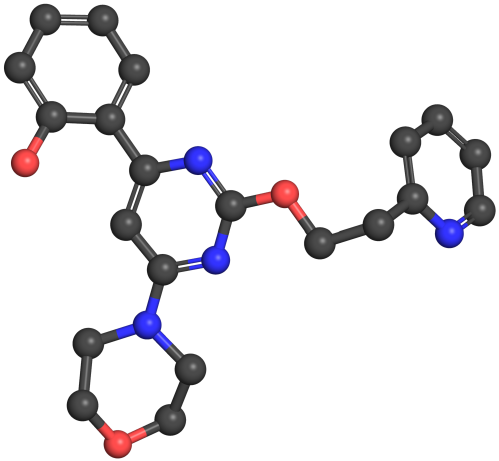}
        \caption{$\mathrm{C_{21}H_{22}N_{4}O_{3}}$}
        \label{fig:A1}
    \end{subfigure}\hfill
    \begin{subfigure}[t]{0.31\linewidth}
        \centering
        \includegraphics[width=\linewidth]{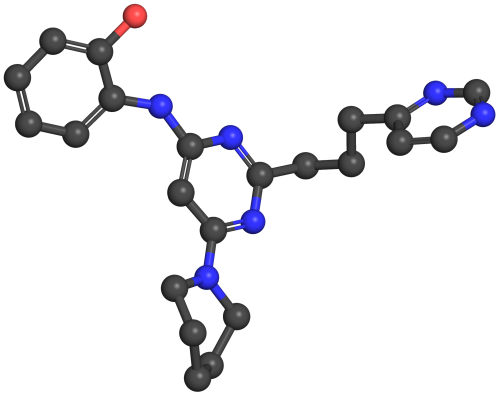}
        \caption{$\mathrm{C_{22}H_{26}N_{6}O}$}
        \label{fig:A2}
    \end{subfigure}\hfill
    \begin{subfigure}[t]{0.31\linewidth}
        \centering
        \includegraphics[width=\linewidth]{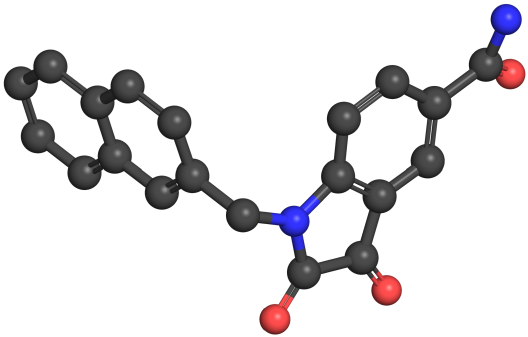}
        \caption{$\mathrm{C_{20}H_{14}N_{2}O_{3}}$}
        \label{fig:A3}
    \end{subfigure}\hfill
    \\[4pt]
    \begin{subfigure}[t]{0.31\linewidth}
        \centering
        \includegraphics[width=\linewidth]{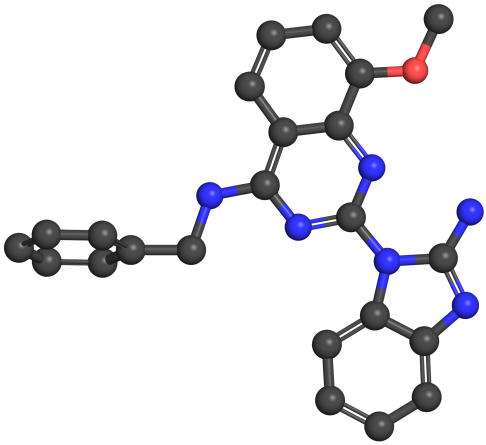}
        \caption{$\mathrm{C_{23}H_{20}N_{6}O}$}
        \label{fig:A4}
    \end{subfigure}\hfill
    \begin{subfigure}[t]{0.31\linewidth}
        \centering
        \includegraphics[width=\linewidth]{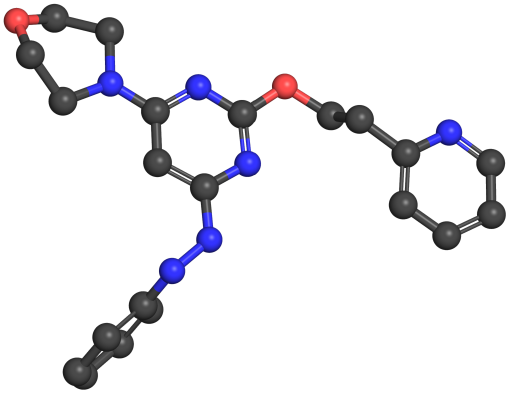}
        \caption{$\mathrm{C_{21}H_{24}N_{6}O_{2}}$}
        \label{fig:A5}
    \end{subfigure}\hfill
    \begin{subfigure}[t]{0.31\linewidth}
        \centering
        \includegraphics[width=\linewidth]{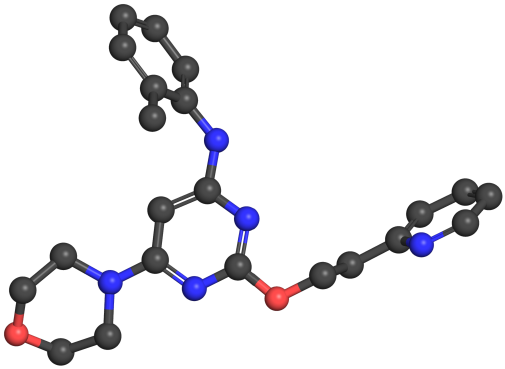}
        \caption{$\mathrm{C_{22}H_{25}N_{5}O_{2}}$}
        \label{fig:A6}
    \end{subfigure}
    \caption{The six best agent-found molecules that Pareto-dominate all the literature seeds over docking score, QED, and SA score. \textbf{a} and \textbf{b} was found by Qwen, \textbf{c} and \textbf{d} by DeepSeek, and \textbf{e} and \textbf{f} by Claude.}
    \label{fig:discoveries}
\end{figure}

\begin{table}[!ht]
\centering
\caption{Compounds and SELFIES strings converted to SMILES strings for the six best literature seeds and the six best molecules found during agentic optimization. SMILES here are RDKit-canonical strings.}
\label{tab:smiles}
\scriptsize
\setlength{\tabcolsep}{3pt}
\newcommand{\smiles}[1]{\texttt{\seqsplit{#1}}}
\begin{tabular}{p{0.18\textwidth} p{0.78\textwidth}}
\toprule
Compound & SMILES \\
\midrule
\multicolumn{2}{l}{\textit{Literature seeds}} \\
MDL-28170 & \smiles{CC(C)[C@H](NC(=O)OCc1ccccc1)C(=O)NC(C=O)Cc1ccccc1} \\
VBY-825 & \smiles{CC[C@H](NC(=O)[C@H](CS(=O)(=O)CC1CC1)N[C@@H](c1ccc(F)cc1)C(F)(F)F)C(=O)C(=O)NC1CC1} \\
Tetrandrine & \smiles{COc1ccc2cc1Oc1ccc(cc1)C[C@H]1c3cc(c(OC)cc3CCN1C)Oc1c(OC)c(OC)cc3c1[C@H](C2)N(C)CC3} \\
YH-1238 & \smiles{COc1cccc2c(N3CCc4ccccc4C3C)nc(Nc3ccc(F)cc3C)nc12} \\
ONO-5334 & \smiles{C[C@@H]1CS/C(=N\textbackslash NC(=O)C(=O)[C@@H](NC(=O)C2CCCCCC2)C2CCOCC2)N1C} \\
Apilimod & \smiles{Cc1cccc(/C=N/Nc2cc(N3CCOCC3)nc(OCCc3ccccn3)n2)c1} \\
\midrule
\multicolumn{2}{l}{\textit{Agents found}} \\
\(\mathrm{C_{21}H_{22}N_{4}O_{3}}\) & \smiles{Oc1ccccc1-c1cc(N2CCOCC2)nc(OCCc2ccccn2)n1} \\
\(\mathrm{C_{22}H_{26}N_{6}O}\) & \smiles{Oc1ccccc1Nc1cc(N2CCCCC2)nc(CCCc2ccncn2)n1} \\
\(\mathrm{C_{20}H_{14}N_{2}O_{3}}\) & \smiles{NC(=O)c1ccc2c(c1)C(=O)C(=O)N2Cc1ccc2ccccc2c1} \\
\(\mathrm{C_{23}H_{20}N_{6}O}\) & \smiles{COc1cccc2c(NCc3ccccc3)nc(-n3c(N)nc4ccccc43)nc12} \\
\(\mathrm{C_{21}H_{24}N_{6}O_{2}}\) & \smiles{c1ccc(NNc2cc(N3CCOCC3)nc(OCCc3ccccn3)n2)cc1} \\
\(\mathrm{C_{22}H_{25}N_{5}O_{2}}\) & \smiles{Cc1ccccc1Nc1cc(N2CCOCC2)nc(OCCc2ccccn2)n1} \\
\bottomrule
\end{tabular}
\end{table}

\begin{table}[!ht]
\centering
\caption{Objective scores and descriptors for the six best literature seeds and the six best molecules found during agentic optimization. Dock is the docking score in kcal\,mol$^{-1}$; QED and SA are from RDKit, and the rest are RDKit descriptors.}
\label{tab:obj_descriptors}
\scriptsize
\setlength{\tabcolsep}{3pt}
\begin{tabular}{lrrrrrrrrr}
\toprule
Compound & Dock & QED & SA & MW & $\log P$ & TPSA & HBD & HBA & Heavy \\
\midrule
\multicolumn{10}{l}{\textit{Literature seeds}} \\
MDL-28170 & $-7.078$ & 0.6533 & 2.760 & 382.46 & 2.864 & 84.50 & 2 & 4 & 28 \\
VBY-825 & $-7.726$ & 0.2624 & 3.782 & 535.56 & 1.955 & 121.44 & 3 & 6 & 36 \\
Tetrandrine & $-8.305$ & 0.2377 & 5.725 & 622.76 & 7.162 & 61.86 & 0 & 8 & 46 \\
YH-1238 & $-8.112$ & 0.4351 & 2.898 & 428.51 & 5.953 & 50.28 & 1 & 5 & 32 \\
ONO-5334 & $-7.619$ & 0.3732 & 3.734 & 438.59 & 1.892 & 100.10 & 2 & 6 & 30 \\
Apilimod & $-8.481$ & 0.4444 & 2.602 & 418.50 & 3.084 & 84.76 & 1 & 8 & 31 \\
\midrule
\multicolumn{10}{l}{\textit{Agents found}} \\
$\mathrm{C_{21}H_{22}N_{4}O_{3}}$ & $-9.146$ & 0.7062 & 2.445 & 378.43 & 2.702 & 80.60 & 1 & 7 & 28 \\
$\mathrm{C_{22}H_{26}N_{6}O}$ & $-9.020$ & 0.5920 & 2.592 & 390.49 & 3.881 & 87.06 & 2 & 7 & 29 \\
$\mathrm{C_{20}H_{14}N_{2}O_{3}}$ & $-8.948$ & 0.7498 & 1.999 & 330.34 & 2.668 & 80.47 & 1 & 3 & 25 \\
$\mathrm{C_{23}H_{20}N_{6}O}$ & $-8.874$ & 0.4633 & 2.304 & 396.45 & 4.172 & 90.88 & 2 & 7 & 30 \\
$\mathrm{C_{21}H_{24}N_{6}O_{2}}$ & $-8.772$ & 0.5664 & 2.515 & 392.46 & 2.769 & 84.43 & 2 & 8 & 29 \\
$\mathrm{C_{22}H_{25}N_{5}O_{2}}$ & $-8.771$ & 0.6618 & 2.404 & 391.48 & 3.382 & 72.40 & 1 & 7 & 29 \\
\bottomrule
\end{tabular}
\end{table}

\clearpage

\begin{figure}[!ht]
    \centering
    \begin{subfigure}[t]{0.48\linewidth}
        \centering
        \includegraphics[width=\linewidth]{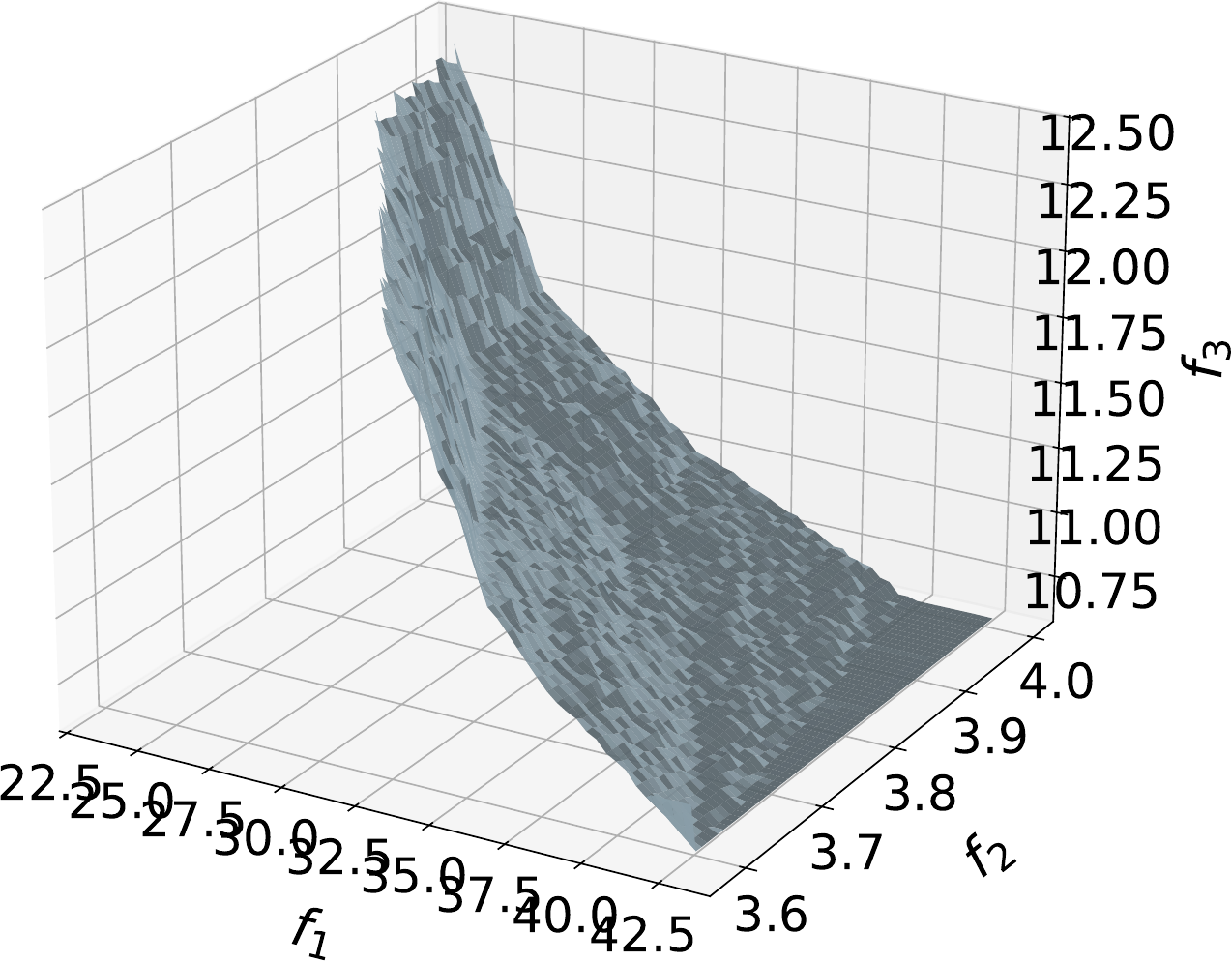}
        \caption{carside}
        \label{fig:carside_attainment}
    \end{subfigure}\hfill
    \begin{subfigure}[t]{0.48\linewidth}
        \centering
        \includegraphics[width=\linewidth]{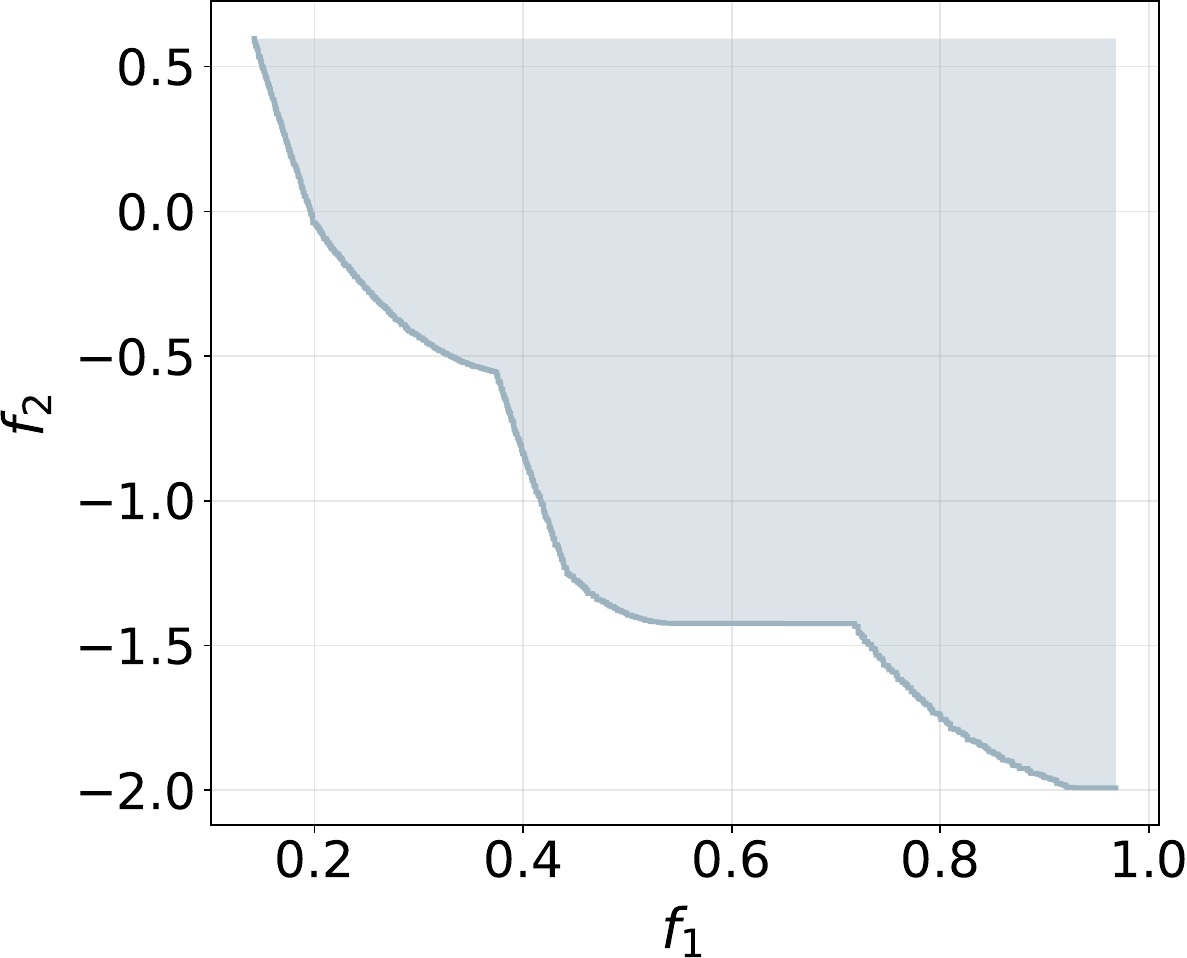}
        \caption{ct1}
        \label{fig:ct1_attainment}
    \end{subfigure}
 
    \vspace{1em}
 
    \begin{subfigure}[t]{0.48\linewidth}
        \centering
        \includegraphics[width=\linewidth]{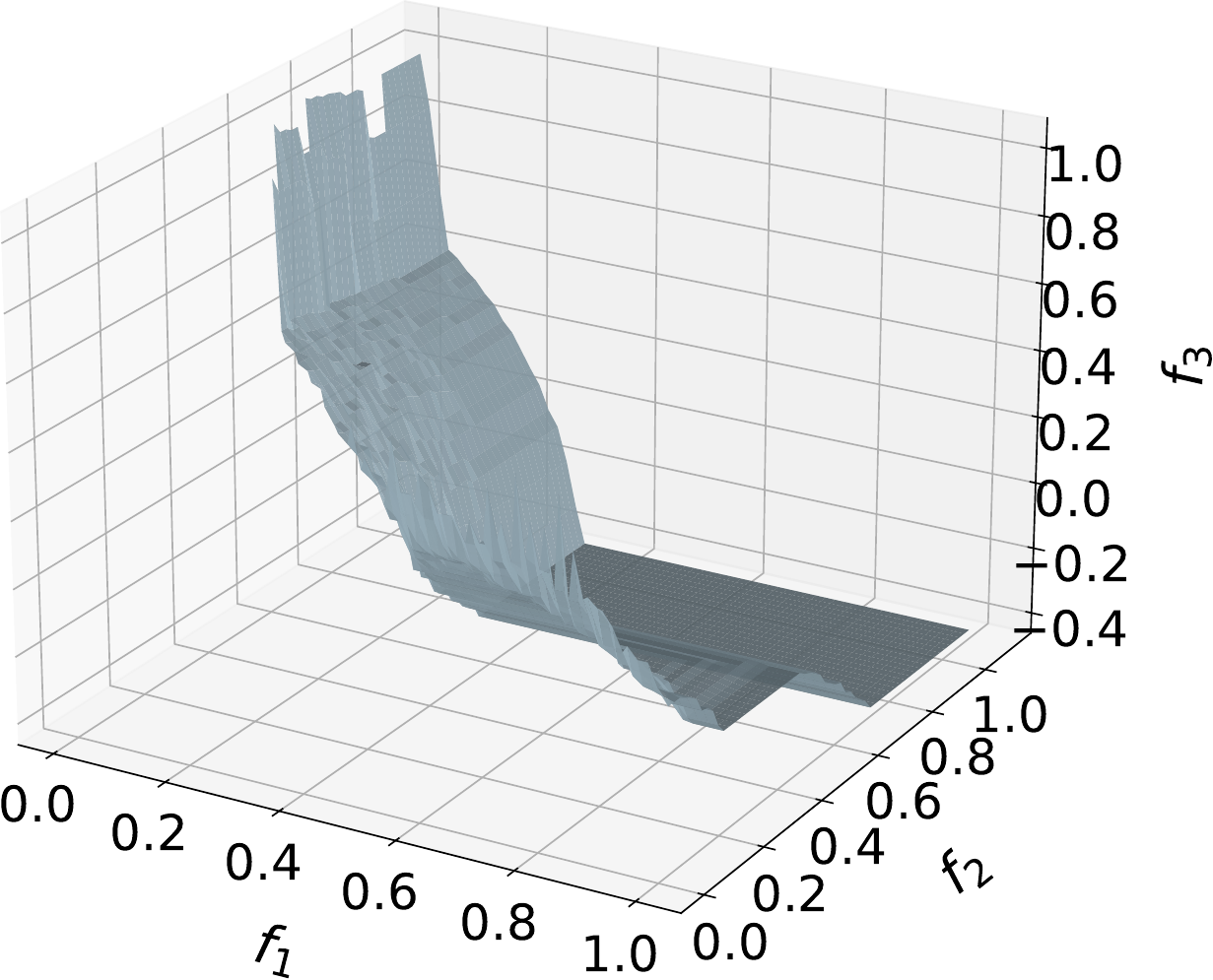}
        \caption{re37}
        \label{fig:re37_attainment}
    \end{subfigure}\hfill
    \begin{subfigure}[t]{0.48\linewidth}
        \centering
        \includegraphics[width=\linewidth]{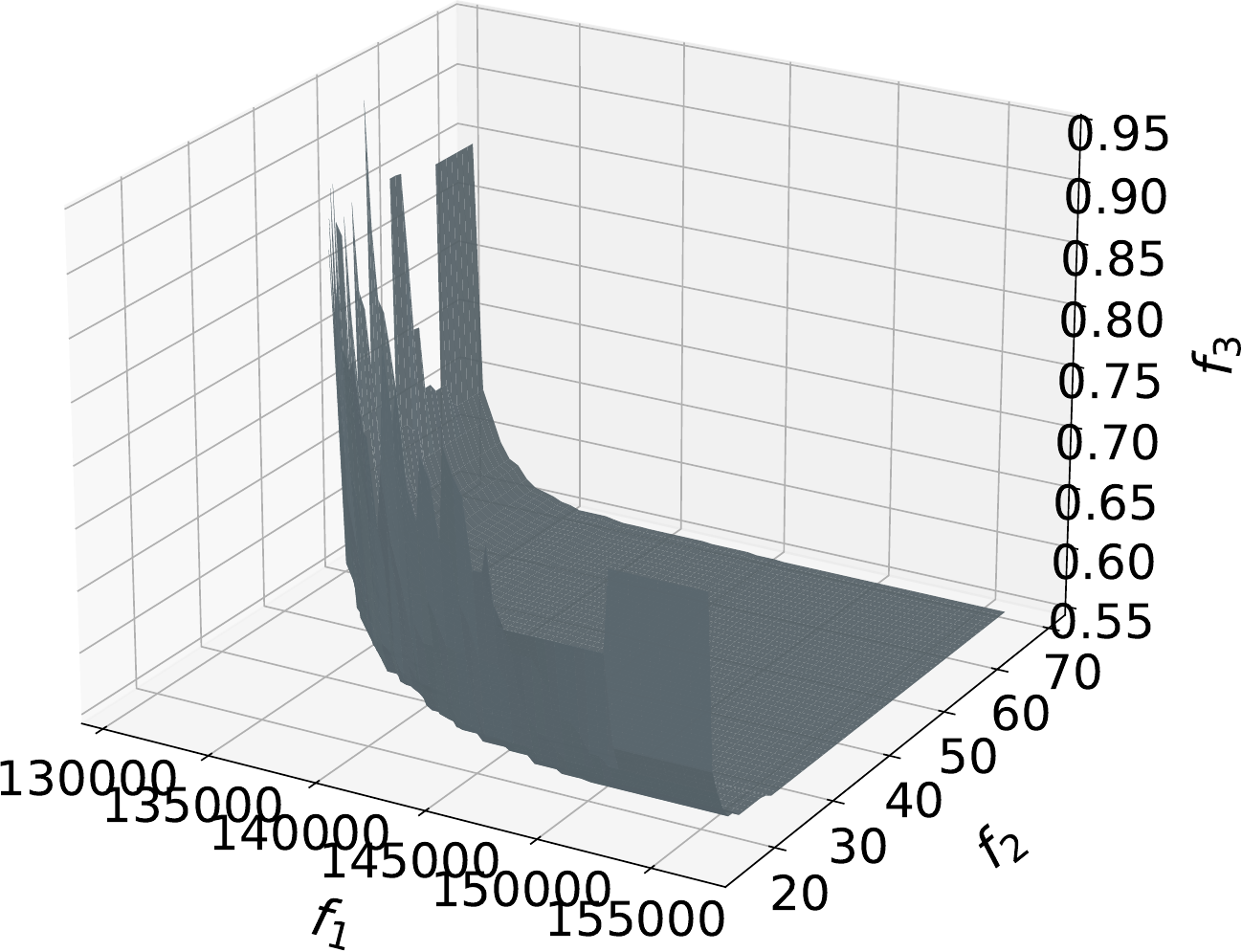}
        \caption{opf118}
        \label{fig:opf118_attainment}
    \end{subfigure}
 
    \vspace{1em}
 
    \begin{minipage}[c]{0.48\linewidth}
        \centering
        \includegraphics[width=\linewidth]{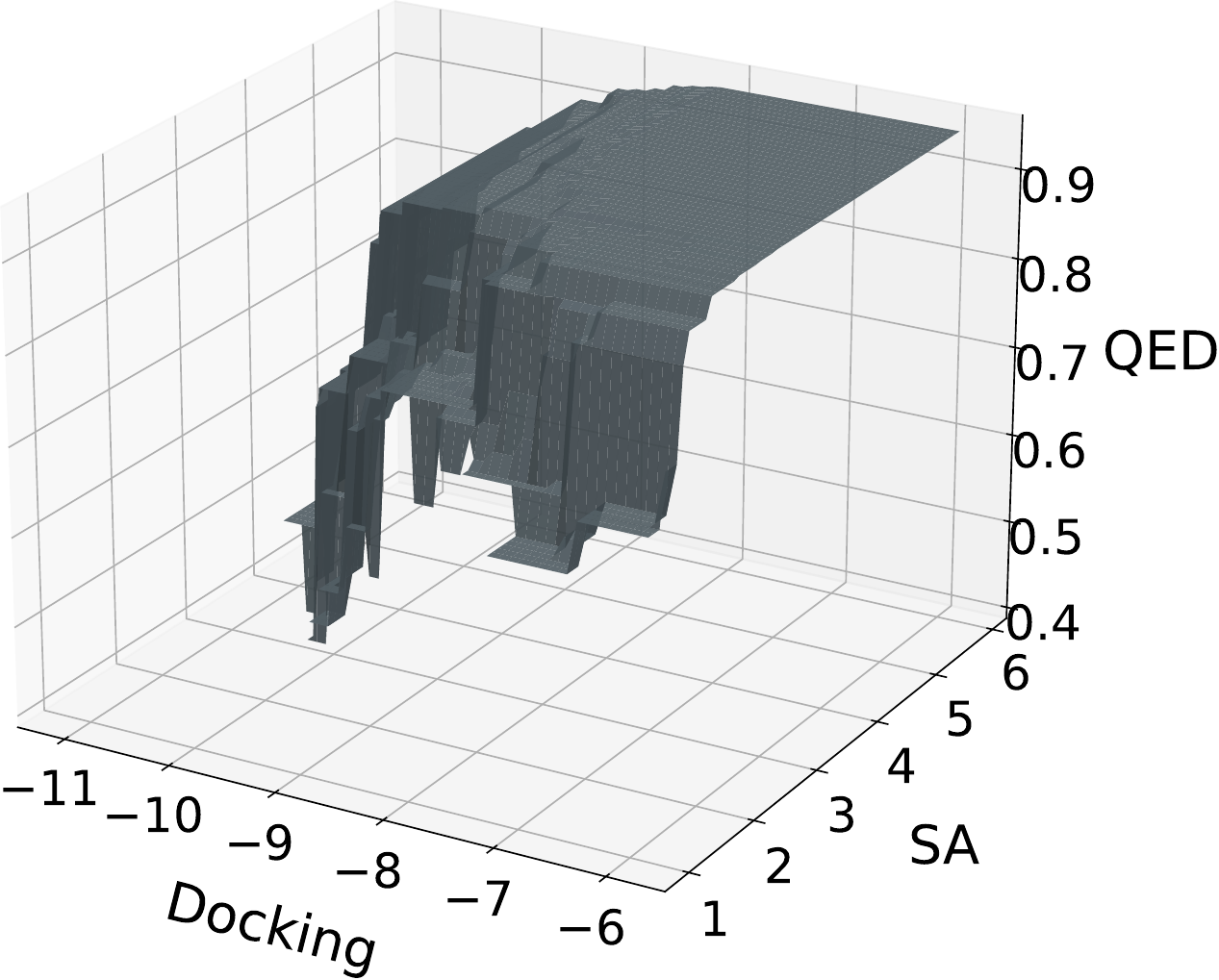}
    \end{minipage}\hfill
    \begin{minipage}[c]{0.48\linewidth}
        \caption{Reference fronts used for IGD$^+$ (Section~\ref{sec:fronts}), shown as attainment surfaces: (a)--(d) the four real-valued benchmarks and, at left, 6W63 docking. ct1 has two objectives, so its front is drawn as a curve.}
        \label{fig:attainment}
    \end{minipage}
\end{figure}

\clearpage

\begin{table}[tp]
    \centering
    \caption{MOEA baseline settings. The four algorithms share generator, variation,
    evaluation interface and budget accounting, and differ only in environmental
    selection, so a difference between them isolates the selection rule. Selection is
    taken from \textit{pymoo}. Variation depends on the genome, not on the algorithm.}
    \begin{tabular}{@{}l p{0.68\linewidth}@{}}
        \toprule
        \multicolumn{2}{@{}l}{\emph{Shared by all}} \\
        \midrule
        Parent selection & binary tournament (constrained rank, crowding distance) \\
        Vector variation & \textit{pymoo} SBX ($\eta_c=15$, $p_c=0.9$) and polynomial mutation ($\eta_m=20$, $p_m=1/n$) \\
        SELFIES variation & token-level mutation; crossover in the encoding, applied with $p=0.5$ \\
        Population & 100 on the vector benchmarks, 50 on docking \\
        \bottomrule
    \end{tabular}
        \label{tab:ea_params}
\end{table}
 
\begin{table}[!ht]
\centering
\caption{Median (first line) and interquartile range (second line, in parentheses) of final HV (higher is better) and best-of-run IGD$^+$ (lower is better) on the four real-valued benchmarks. Baselines have 30 runs and agents 10 runs on every benchmark. The best median in each column is in bold.}
\label{tab:app_synth_medians}
\scriptsize
\setlength{\tabcolsep}{3pt}
\begin{tabular}{lcccccccc}
\toprule
 & \multicolumn{2}{c}{carside} & \multicolumn{2}{c}{ct1} & \multicolumn{2}{c}{re37} & \multicolumn{2}{c}{opf118} \\
\cmidrule(lr){2-3}\cmidrule(lr){4-5}\cmidrule(lr){6-7}\cmidrule(lr){8-9}
Configuration & HV & IGD$^+$ & HV & IGD$^+$ & HV & IGD$^+$ & HV & IGD$^+$ \\
\midrule
SMS-EMOA & 0.518 & 0.0122 & 0.503 & 0.1406 & 0.648 & 0.0090 & 0.694 & 0.0147 \\
 & (0.007) & (0.0019) & (0.060) & (0.0381) & (0.001) & (0.0009) & (0.011) & (0.0064) \\[1pt]
AGE-MOEA-II & 0.498 & 0.0204 & 0.531 & 0.1207 & 0.644 & 0.0095 & 0.657 & 0.0311 \\
 & (0.009) & (0.0031) & (0.045) & (0.0377) & (0.001) & (0.0007) & (0.014) & (0.0048) \\[1pt]
NSGA-II & 0.487 & 0.0255 & 0.521 & 0.1276 & 0.640 & 0.0125 & 0.659 & 0.0344 \\
 & (0.012) & (0.0061) & (0.067) & (0.0494) & (0.002) & (0.0014) & (0.016) & (0.0063) \\[1pt]
SPEA2 & 0.485 & 0.0237 & 0.517 & 0.1370 & 0.638 & 0.0102 & 0.659 & 0.0232 \\
 & (0.018) & (0.0036) & (0.042) & (0.0260) & (0.004) & (0.0008) & (0.019) & (0.0074) \\[1pt]
\midrule
Qwen & \textbf{0.526} & \textbf{0.0105} & 0.520 & 0.1258 & 0.642 & 0.0102 & \textbf{0.714} & \textbf{0.0069} \\
 & (0.006) & (0.0023) & (0.031) & (0.0243) & (0.012) & (0.0022) & (0.007) & (0.0030) \\[1pt]
DeepSeek & 0.508 & 0.0152 & 0.549 & 0.1039 & 0.652 & \textbf{0.0088} & 0.676 & 0.0164 \\
 & (0.008) & (0.0025) & (0.049) & (0.0272) & (0.002) & (0.0012) & (0.023) & (0.0063) \\[1pt]
GPT & 0.506 & 0.0200 & \textbf{0.612} & \textbf{0.0751} & \textbf{0.653} & 0.0101 & 0.672 & 0.0196 \\
 & (0.006) & (0.0020) & (0.021) & (0.0135) & (0.004) & (0.0015) & (0.014) & (0.0101) \\[1pt]
Gemini & 0.501 & 0.0207 & 0.530 & 0.1170 & 0.651 & 0.0091 & 0.674 & 0.0270 \\
 & (0.009) & (0.0039) & (0.062) & (0.0439) & (0.005) & (0.0010) & (0.010) & (0.0085) \\[1pt]
Claude & 0.502 & 0.0221 & 0.543 & 0.1098 & 0.648 & 0.0099 & 0.671 & 0.0194 \\
 & (0.013) & (0.0043) & (0.096) & (0.0504) & (0.002) & (0.0004) & (0.009) & (0.0026) \\[1pt]
\bottomrule
\end{tabular}
\end{table}
 
\begin{table}[!ht]
\centering
\caption{Median and interquartile range (in parentheses) of final HV and best-of-run IGD$^+$ on 6W63 docking in both conditions, with the number of runs $n$. Both conditions share one normalization and reference set. A dash marks a model with no completed runs (refusals). The best median in each column is in bold.}
\label{tab:app_dock_medians}
\scriptsize
\setlength{\tabcolsep}{3pt}
\begin{tabular}{lrccrcc}
\toprule
 & \multicolumn{3}{c}{Random initialization} & \multicolumn{3}{c}{Literature seeded} \\
\cmidrule(lr){2-4}\cmidrule(lr){5-7}
Configuration & $n$ & HV & IGD$^+$ & $n$ & HV & IGD$^+$ \\
\midrule
SMS-EMOA & 25 & 0.389 (0.023) & 0.2998 (0.0569) & 25 & 0.673 (0.032) & 0.0552 (0.0120) \\
AGE-MOEA-II & 25 & 0.368 (0.037) & 0.3240 (0.0663) & 25 & 0.668 (0.021) & 0.0554 (0.0148) \\
NSGA-II & 25 & 0.387 (0.035) & 0.2757 (0.0634) & 25 & 0.663 (0.022) & 0.0621 (0.0126) \\
SPEA2 & 25 & 0.368 (0.030) & 0.3206 (0.0430) & 25 & 0.649 (0.020) & 0.0657 (0.0096) \\
\midrule
Qwen & 10 & 0.409 (0.024) & 0.2786 (0.0258) & 10 & \textbf{0.700} (0.023) & 0.0527 (0.0179) \\
DeepSeek & 10 & \textbf{0.423} (0.059) & \textbf{0.2727} (0.0567) & 10 & 0.681 (0.020) & \textbf{0.0515} (0.0105) \\
GPT & 4 & 0.407 (0.041) & 0.3052 (0.0345) & 0 & -- & -- \\
Gemini & 0 & -- & -- & 0 & -- & -- \\
Claude & 4 & 0.363 (0.023) & 0.3473 (0.0708) & 5 & 0.690 (0.015) & 0.0528 (0.0044) \\
\bottomrule
\end{tabular}
\end{table}
 
\begin{table}[!ht]
\centering
\caption{All pairwise comparisons. Each entry is the Vargha--Delaney $\hat{A}_{12}$ of the row configuration against the column configuration, the probability that a run of the row configuration is better, so values above .5 favor the row. Entries above the diagonal are for HV and entries below it for IGD$^+$. Bold entries are significant (Mann--Whitney U, $p<0.05$ after Holm--Bonferroni correction over all pairs for that problem and indicator). Abbreviations: SMS = SMS-EMOA, AGE = AGE-MOEA-II, NSGA = NSGA-II, SPEA = SPEA2, Qw = Qwen, DS = DeepSeek, Gem = Gemini, Cl = Claude.}
\label{tab:app_pairwise}
\scriptsize
\setlength{\tabcolsep}{1.0pt}
\begin{minipage}[t]{0.49\textwidth}\centering
\textbf{carside}\\[2pt]
\begin{tabular}{lccccccccc}
 & {\tiny SMS} & {\tiny AGE} & {\tiny NSGA} & {\tiny SPEA} & {\tiny Qw} & {\tiny DS} & {\tiny GPT} & {\tiny Gem} & {\tiny Cl} \\
\midrule
SMS & -- & \textbf{.99} & \textbf{1} & \textbf{1} & \textbf{.06} & \textbf{.87} & \textbf{.92} & \textbf{.97} & \textbf{.92} \\
AGE & \textbf{.00} & -- & \textbf{.79} & \textbf{.80} & \textbf{.00} & \textbf{.14} & \textbf{.14} & .41 & .36 \\
NSGA & \textbf{.00} & \textbf{.13} & -- & .55 & \textbf{.00} & \textbf{.03} & \textbf{.00} & \textbf{.12} & \textbf{.13} \\
SPEA & \textbf{.00} & \textbf{.19} & .61 & -- & \textbf{.00} & \textbf{.03} & \textbf{.00} & \textbf{.12} & \textbf{.12} \\
Qw & \textbf{.82} & \textbf{1} & \textbf{1} & \textbf{1} & -- & \textbf{1} & \textbf{1} & \textbf{1} & \textbf{1} \\
DS & \textbf{.14} & \textbf{.94} & \textbf{1} & \textbf{1} & \textbf{.02} & -- & .57 & .79 & .69 \\
GPT & \textbf{.00} & .53 & \textbf{.93} & \textbf{.90} & \textbf{.00} & \textbf{.05} & -- & .76 & .67 \\
Gem & \textbf{.00} & .45 & \textbf{.81} & .72 & \textbf{.00} & \textbf{.06} & .39 & -- & .44 \\
Cl & \textbf{.02} & .45 & .80 & .72 & \textbf{.00} & \textbf{.11} & .39 & .53 & -- \\
\end{tabular}
\end{minipage}
\hfill
\begin{minipage}[t]{0.49\textwidth}\centering
\textbf{ct1}\\[2pt]
\begin{tabular}{lccccccccc}
 & {\tiny SMS} & {\tiny AGE} & {\tiny NSGA} & {\tiny SPEA} & {\tiny Qw} & {\tiny DS} & {\tiny GPT} & {\tiny Gem} & {\tiny Cl} \\
\midrule
SMS & -- & .39 & .40 & .46 & .43 & .20 & \textbf{.00} & .40 & .23 \\
AGE & .62 & -- & .52 & .61 & .57 & .29 & \textbf{.00} & .50 & .33 \\
NSGA & .57 & .46 & -- & .56 & .51 & .28 & \textbf{.01} & .49 & .29 \\
SPEA & .51 & .37 & .43 & -- & .48 & .20 & \textbf{.00} & .42 & .24 \\
Qw & .60 & .48 & .55 & .62 & -- & .26 & .10 & .44 & .29 \\
DS & .83 & .76 & .76 & .83 & .73 & -- & .20 & .67 & .51 \\
GPT & \textbf{1} & \textbf{.99} & \textbf{.98} & \textbf{1} & .90 & .78 & -- & .90 & .73 \\
Gem & .65 & .55 & .58 & .66 & .55 & .33 & .11 & -- & .36 \\
Cl & .81 & .72 & .77 & \textbf{.85} & .75 & .49 & .33 & .66 & -- \\
\end{tabular}
\end{minipage}
\\[8pt]
\begin{minipage}[t]{0.49\textwidth}\centering
\textbf{re37}\\[2pt]
\begin{tabular}{lccccccccc}
 & {\tiny SMS} & {\tiny AGE} & {\tiny NSGA} & {\tiny SPEA} & {\tiny Qw} & {\tiny DS} & {\tiny GPT} & {\tiny Gem} & {\tiny Cl} \\
\midrule
SMS & -- & \textbf{.92} & \textbf{1} & \textbf{1} & \textbf{.86} & \textbf{.12} & \textbf{.01} & .21 & .58 \\
AGE & .31 & -- & \textbf{.98} & \textbf{1} & .50 & \textbf{.00} & \textbf{.00} & \textbf{.00} & .19 \\
NSGA & \textbf{.01} & \textbf{.01} & -- & \textbf{.84} & .46 & \textbf{.00} & \textbf{.00} & \textbf{.00} & \textbf{.09} \\
SPEA & \textbf{.05} & \textbf{.13} & \textbf{.92} & -- & .37 & \textbf{.00} & \textbf{.00} & \textbf{.00} & \textbf{.07} \\
Qw & \textbf{.14} & .22 & .72 & .47 & -- & \textbf{.02} & \textbf{.00} & \textbf{.05} & .23 \\
DS & .54 & .69 & \textbf{.97} & .82 & .79 & -- & .44 & .56 & .88 \\
GPT & .21 & .35 & \textbf{.90} & .58 & .61 & .25 & -- & .69 & \textbf{.96} \\
Gem & .49 & .66 & \textbf{.99} & \textbf{.93} & .85 & .47 & .78 & -- & .83 \\
Cl & \textbf{.12} & .31 & \textbf{.98} & .73 & .65 & .26 & .54 & .16 & -- \\
\end{tabular}
\end{minipage}
\hfill
\begin{minipage}[t]{0.49\textwidth}\centering
\textbf{opf118}\\[2pt]
\begin{tabular}{lccccccccc}
 & {\tiny SMS} & {\tiny AGE} & {\tiny NSGA} & {\tiny SPEA} & {\tiny Qw} & {\tiny DS} & {\tiny GPT} & {\tiny Gem} & {\tiny Cl} \\
\midrule
SMS & -- & \textbf{.99} & \textbf{.98} & \textbf{.98} & \textbf{.04} & .76 & \textbf{.84} & \textbf{.84} & \textbf{.96} \\
AGE & \textbf{.00} & -- & .45 & .50 & \textbf{.00} & \textbf{.12} & \textbf{.08} & \textbf{.09} & \textbf{.10} \\
NSGA & \textbf{.00} & \textbf{.24} & -- & .54 & \textbf{.00} & \textbf{.16} & \textbf{.14} & \textbf{.13} & \textbf{.17} \\
SPEA & \textbf{.07} & \textbf{.84} & \textbf{.93} & -- & \textbf{.00} & \textbf{.14} & \textbf{.11} & \textbf{.10} & \textbf{.13} \\
Qw & \textbf{.92} & \textbf{1} & \textbf{1} & \textbf{1} & -- & \textbf{.99} & \textbf{.99} & \textbf{.90} & \textbf{1} \\
DS & .41 & \textbf{1} & \textbf{1} & \textbf{.91} & \textbf{.06} & -- & .52 & .47 & .59 \\
GPT & .20 & \textbf{.85} & \textbf{.94} & .64 & \textbf{.00} & .26 & -- & .52 & .63 \\
Gem & \textbf{.18} & \textbf{.82} & \textbf{.97} & .46 & \textbf{.04} & .22 & .42 & -- & .58 \\
Cl & \textbf{.18} & \textbf{.96} & \textbf{1} & .74 & \textbf{.00} & .26 & .50 & .66 & -- \\
\end{tabular}
\end{minipage}
\\[8pt]
\begin{minipage}[t]{0.49\textwidth}\centering
\textbf{6W63, random init.}\\[2pt]
\begin{tabular}{lcccccccc}
 & {\tiny SMS} & {\tiny AGE} & {\tiny NSGA} & {\tiny SPEA} & {\tiny Qw} & {\tiny DS} & {\tiny GPT} & {\tiny Cl} \\
\midrule
SMS & -- & .69 & .52 & .68 & .26 & .25 & .29 & .68 \\
AGE & .29 & -- & .36 & .47 & .23 & .19 & .22 & .51 \\
NSGA & .54 & .71 & -- & .63 & .28 & .26 & .27 & .64 \\
SPEA & .33 & .54 & .32 & -- & .19 & .19 & .19 & .54 \\
Qw & .56 & .75 & .48 & .71 & -- & .42 & .47 & .70 \\
DS & .65 & .78 & .60 & .75 & .57 & -- & .65 & .80 \\
GPT & .51 & .68 & .47 & .70 & .40 & .38 & -- & .81 \\
Cl & .22 & .40 & .23 & .40 & .15 & .17 & .19 & -- \\
\end{tabular}
\end{minipage}
\hfill
\begin{minipage}[t]{0.49\textwidth}\centering
\textbf{6W63, lit. seeded}\\[2pt]
\begin{tabular}{lccccccc}
 & {\tiny SMS} & {\tiny AGE} & {\tiny NSGA} & {\tiny SPEA} & {\tiny Qw} & {\tiny DS} & {\tiny Cl} \\
\midrule
SMS & -- & .55 & .63 & \textbf{.83} & \textbf{.12} & .32 & .27 \\
AGE & .45 & -- & .54 & \textbf{.78} & \textbf{.11} & .23 & .21 \\
NSGA & .27 & .34 & -- & \textbf{.77} & \textbf{.00} & \textbf{.14} & \textbf{.10} \\
SPEA & \textbf{.16} & .25 & .38 & -- & \textbf{.00} & \textbf{.02} & \textbf{.00} \\
Qw & .47 & .51 & .65 & .74 & -- & .76 & .80 \\
DS & .54 & .61 & .80 & \textbf{.88} & .54 & -- & .50 \\
Cl & .63 & .67 & .86 & \textbf{.95} & .58 & .56 & -- \\
\end{tabular}
\end{minipage}
\end{table}

\clearpage
 
\begin{figure}[!ht]
    \centering
    \includegraphics[width=0.24\linewidth]{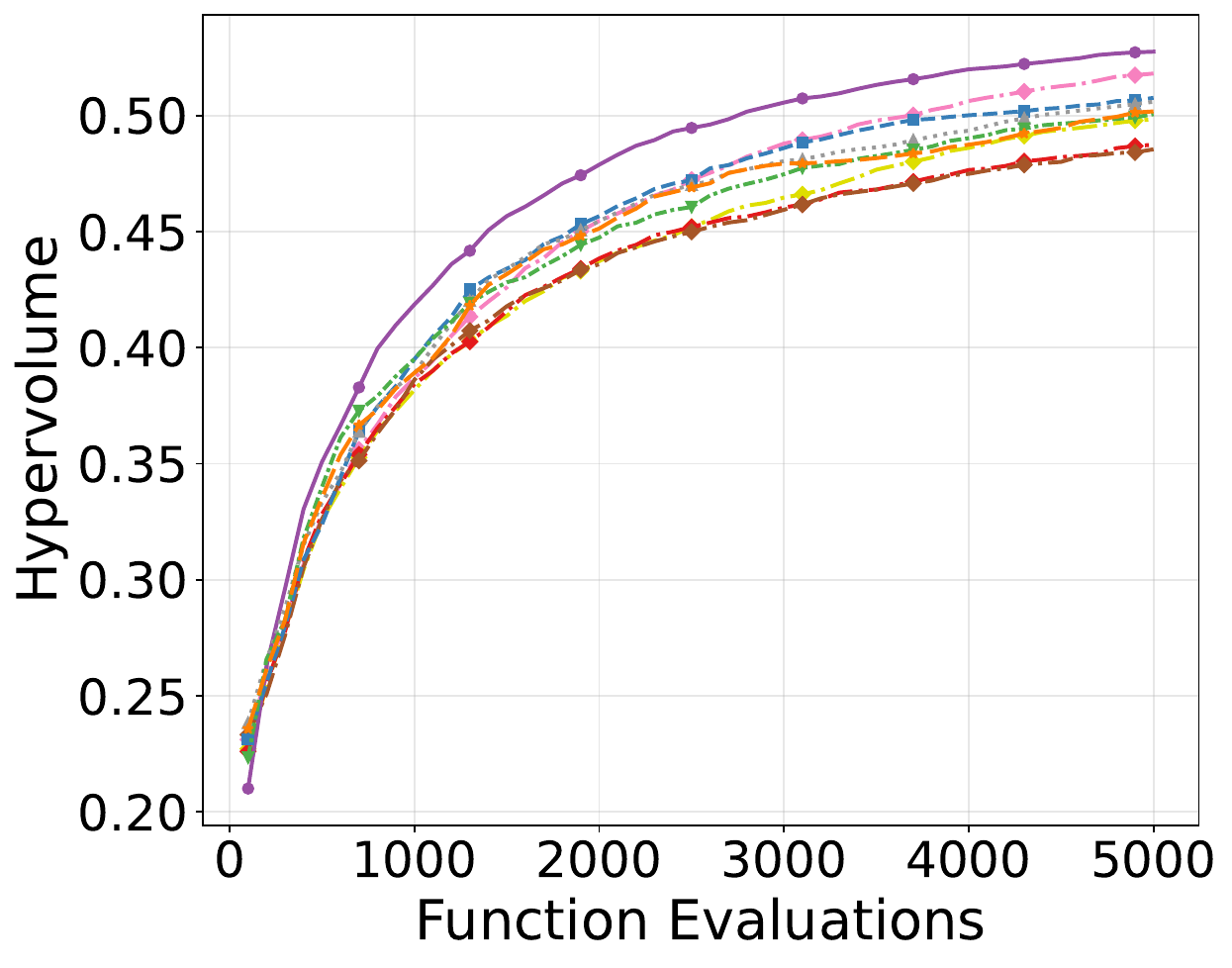}\hfill
    \includegraphics[width=0.24\linewidth]{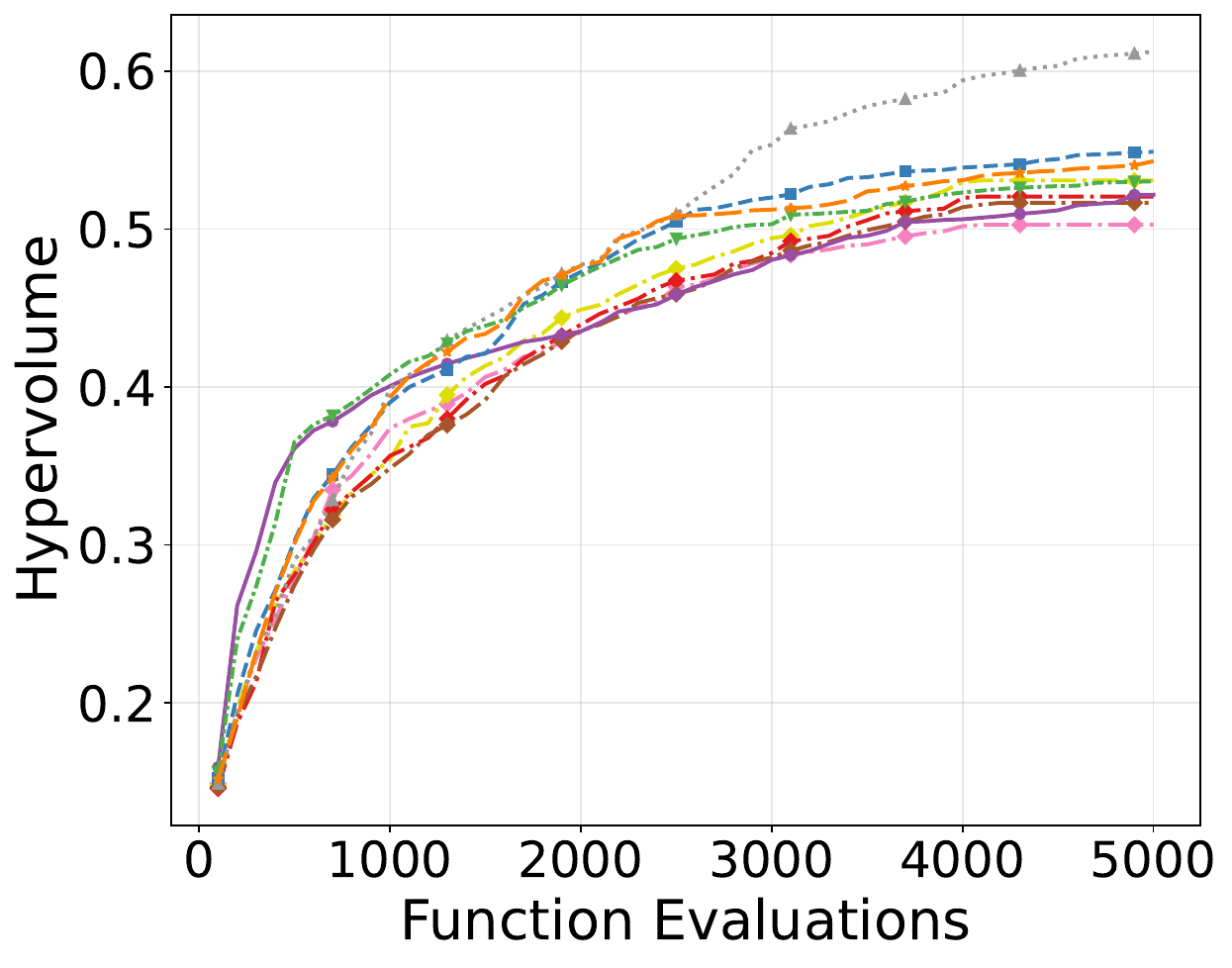}\hfill
    \includegraphics[width=0.24\linewidth]{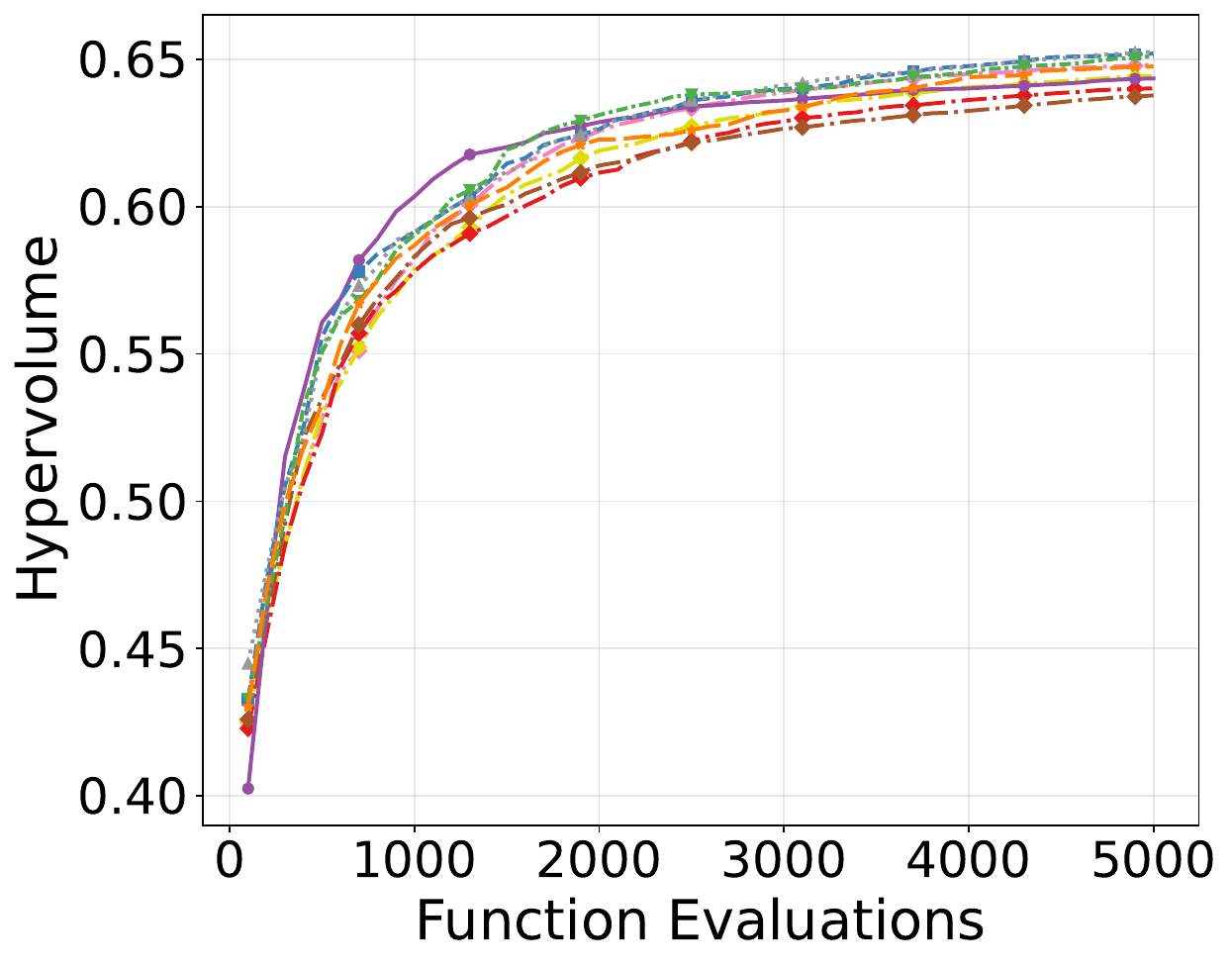}\hfill
    \includegraphics[width=0.24\linewidth]{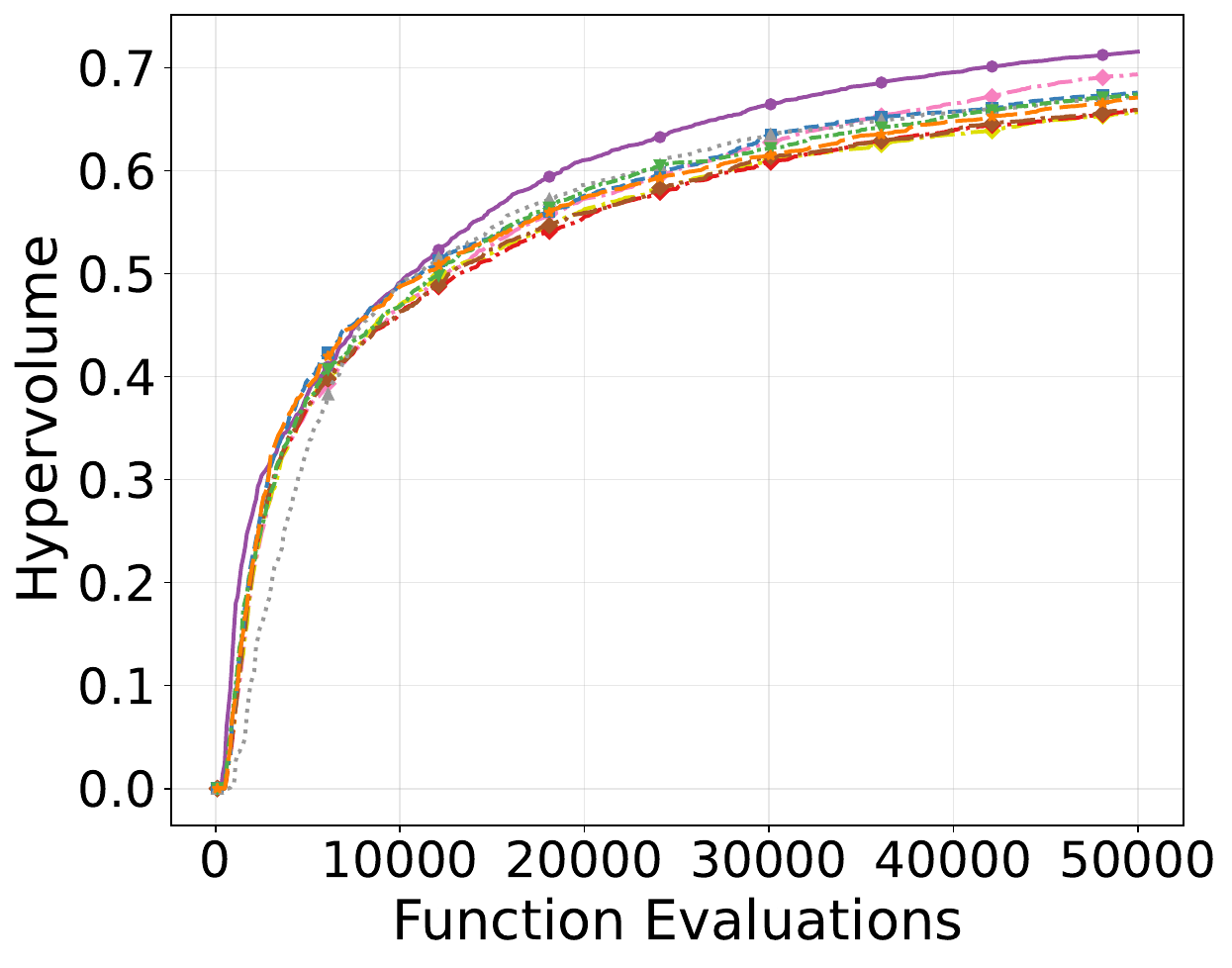}
    \\[4pt]
    \begin{subfigure}[t]{0.24\linewidth}
        \centering
        \includegraphics[width=\linewidth]{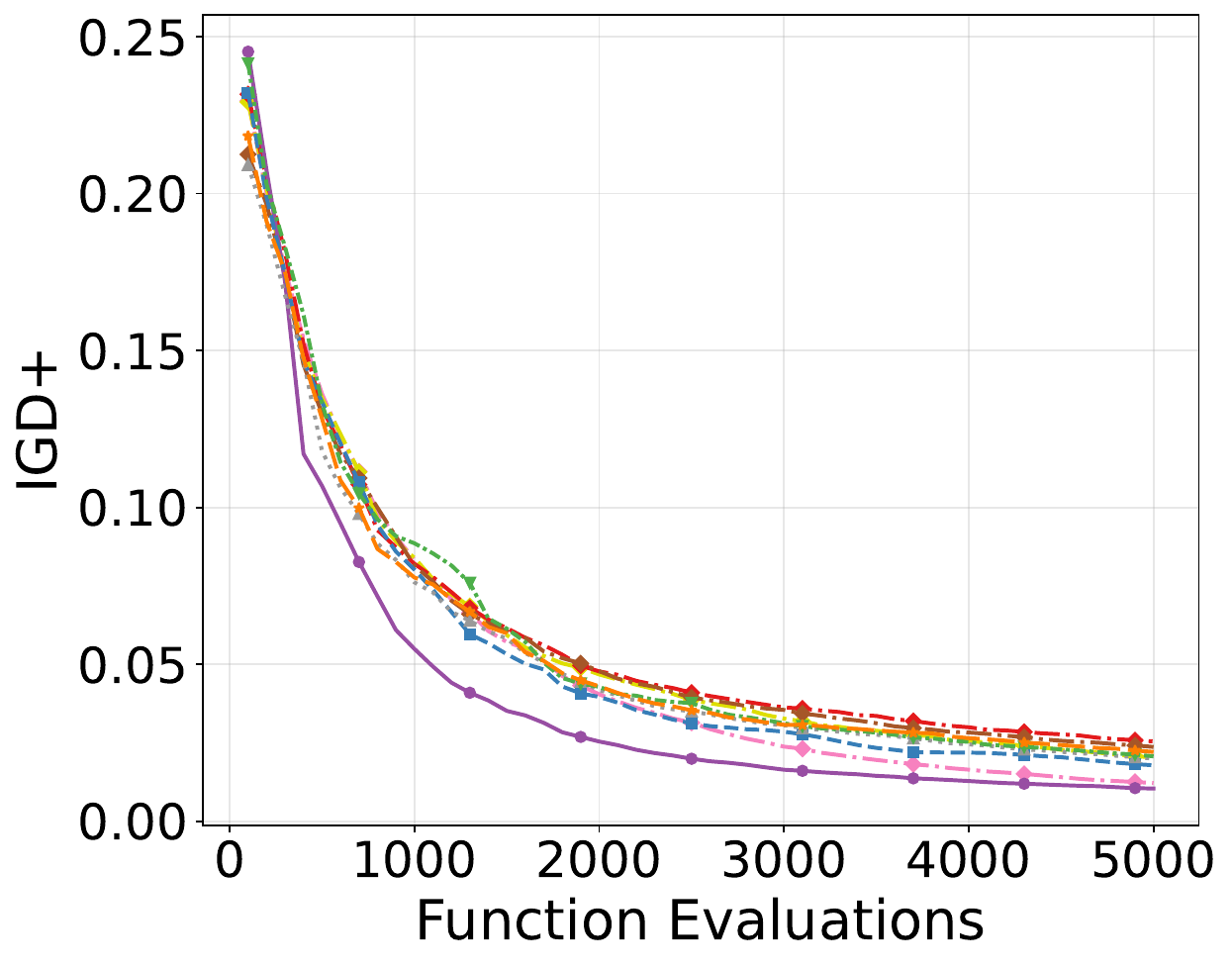}
        \caption{carside}
        \label{fig:carside_curve}
    \end{subfigure}\hfill
    \begin{subfigure}[t]{0.24\linewidth}
        \centering
        \includegraphics[width=\linewidth]{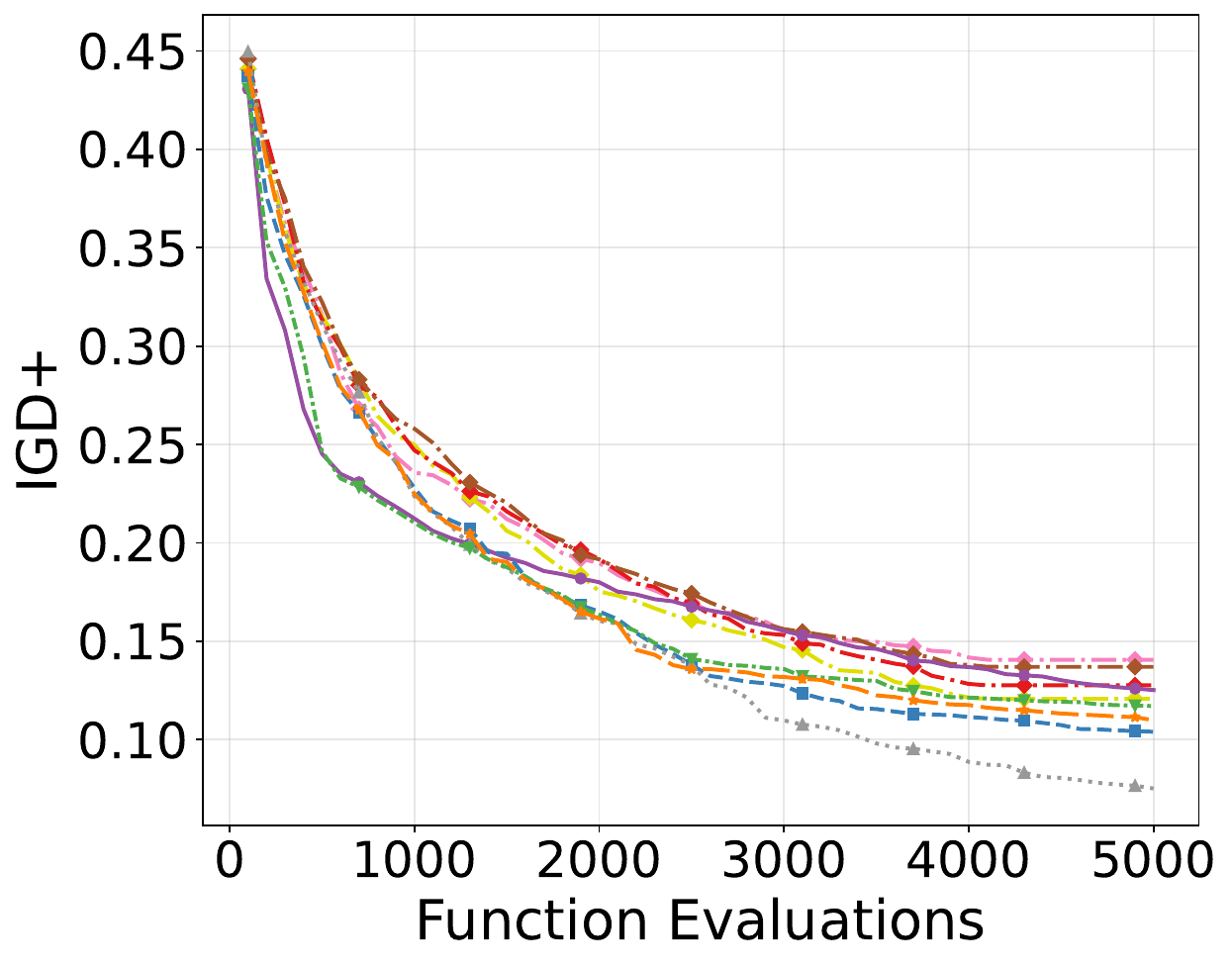}
        \caption{ct1}
        \label{fig:ct1_curve}
    \end{subfigure}\hfill
    \begin{subfigure}[t]{0.24\linewidth}
        \centering
        \includegraphics[width=\linewidth]{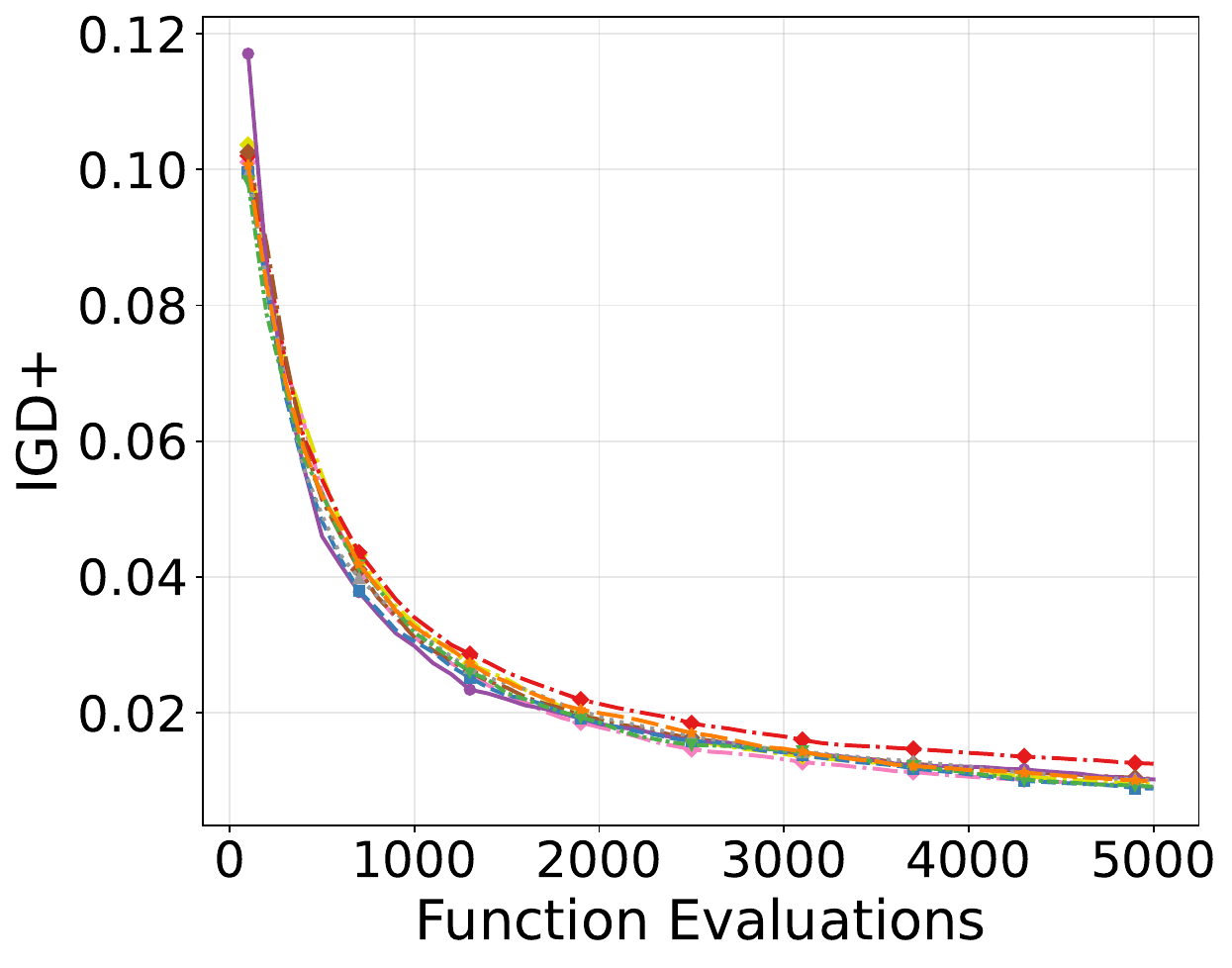}
        \caption{re37}
        \label{fig:re37_curve}
    \end{subfigure}\hfill
    \begin{subfigure}[t]{0.24\linewidth}
        \centering
        \includegraphics[width=\linewidth]{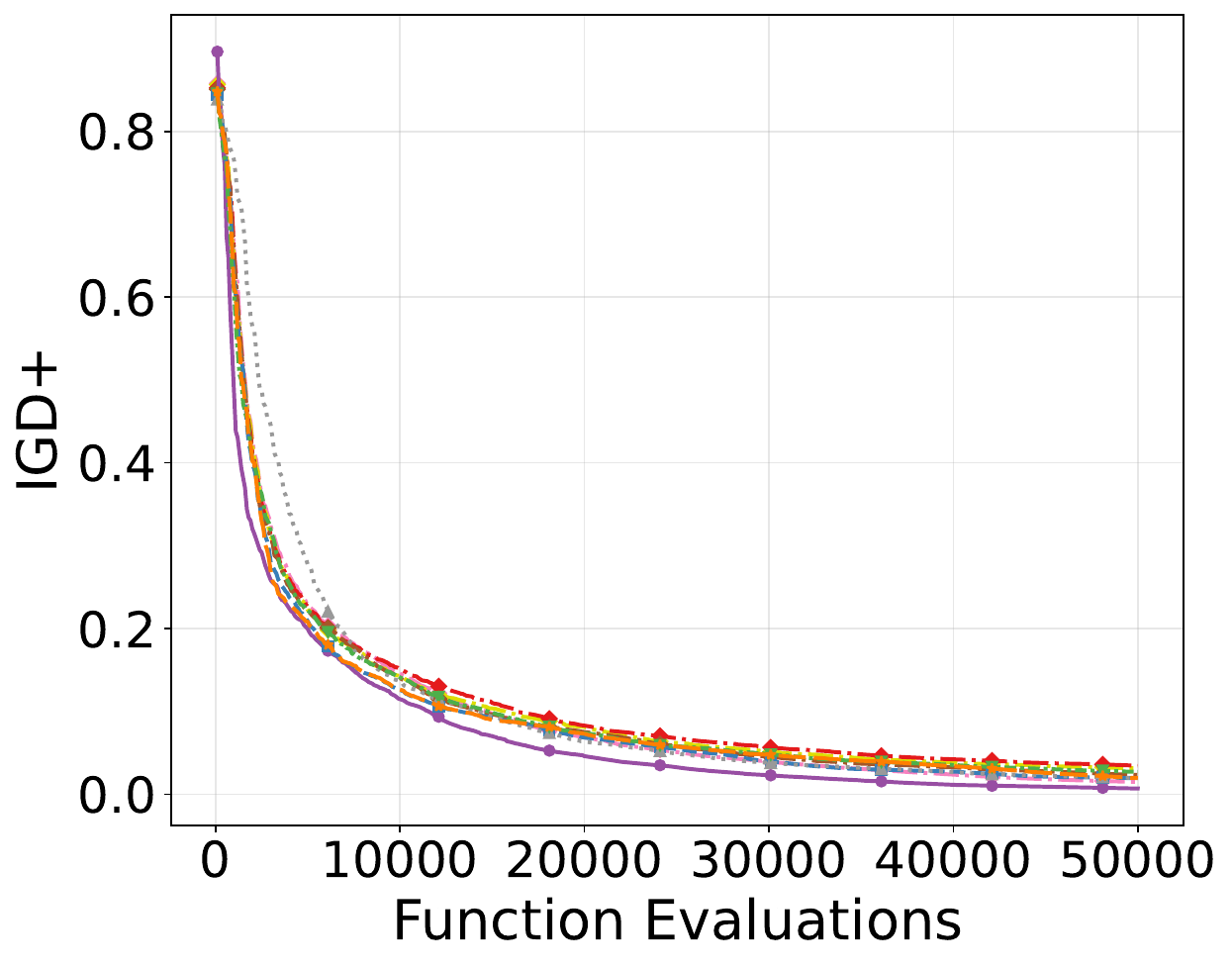}
        \caption{opf118}
        \label{fig:opf118_curve}
    \end{subfigure}
    \caption{Convergence on the four real-valued benchmarks (a--d): HV (top row, higher is better) and IGD$^+$ (bottom row, lower is better) against function evaluations. Each line is the median over runs for one configuration; colors match Fig.~\ref{fig:tradeoffs}. Vertical scales differ between panels.}
    \label{fig:synthetics_curves}
\end{figure}
 
\begin{figure}[!ht]
    \centering
    \begin{subfigure}[t]{0.24\linewidth}
        \centering
        \includegraphics[width=\linewidth]{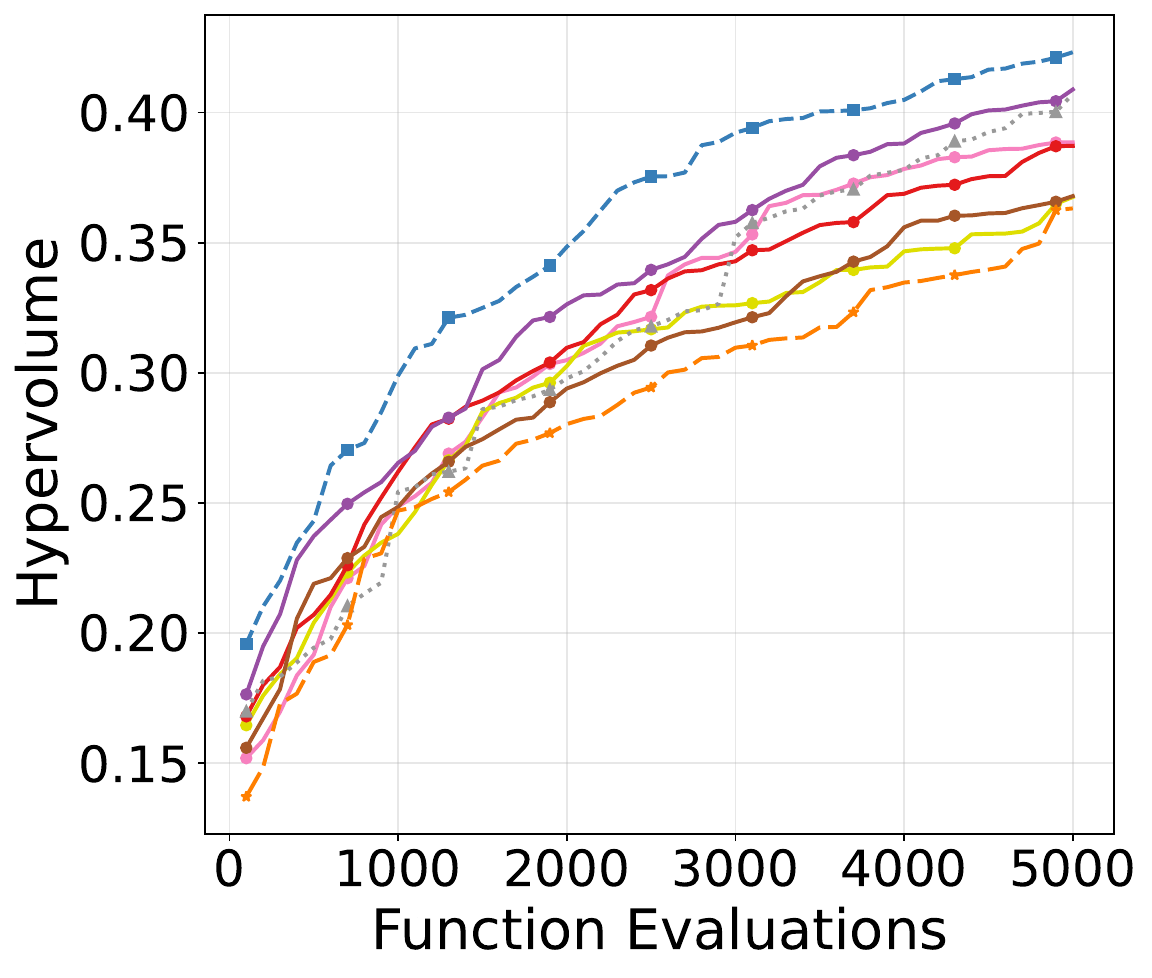}
        \caption{HV, rand. init.}
        \label{fig:docking_rand_hv_curve}
    \end{subfigure}\hfill
    \begin{subfigure}[t]{0.24\linewidth}
        \centering
        \includegraphics[width=\linewidth]{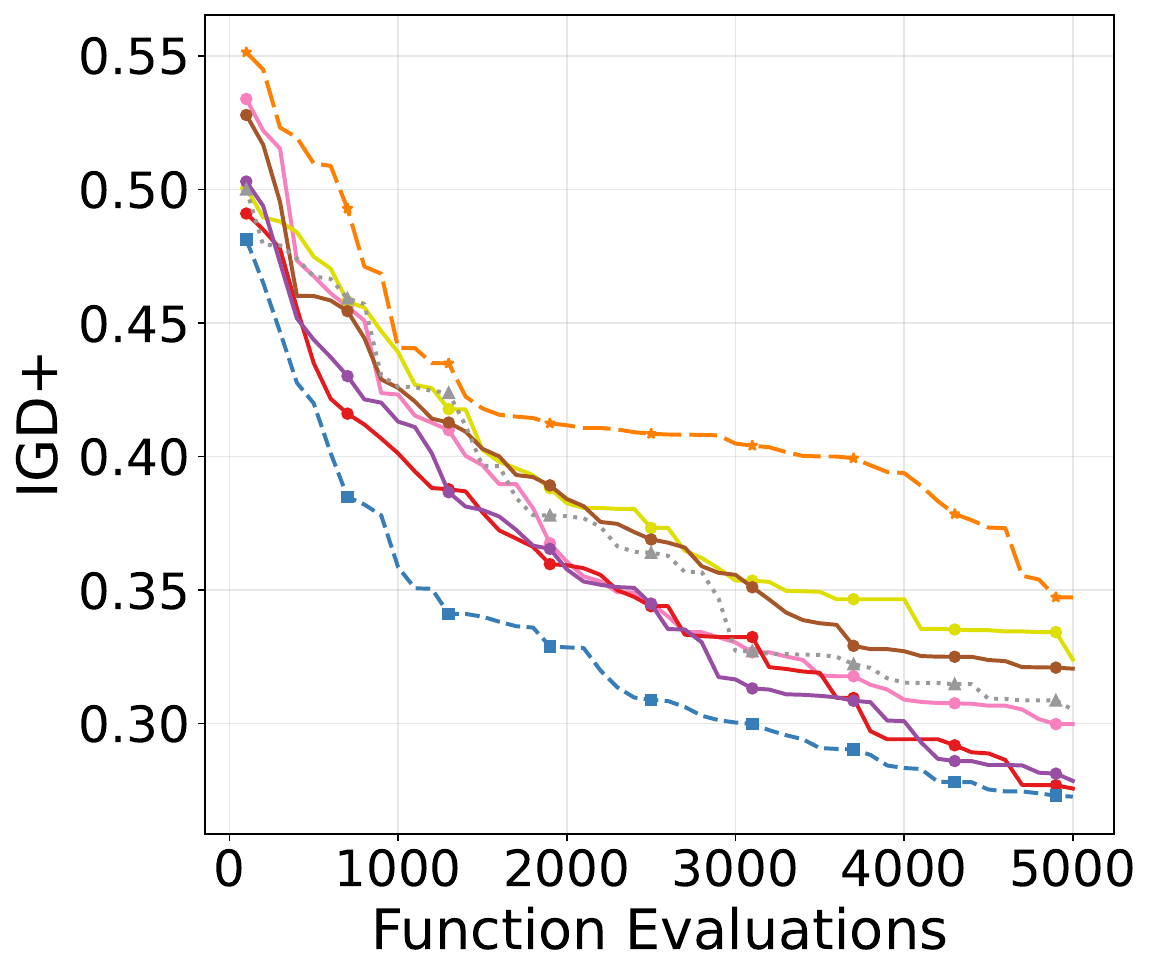}
        \caption{IGD$^+$, rand. init.}
        \label{fig:docking_rand_igdplus_curve}
    \end{subfigure}\hfill
    \begin{subfigure}[t]{0.24\linewidth}
        \centering
        \includegraphics[width=\linewidth]{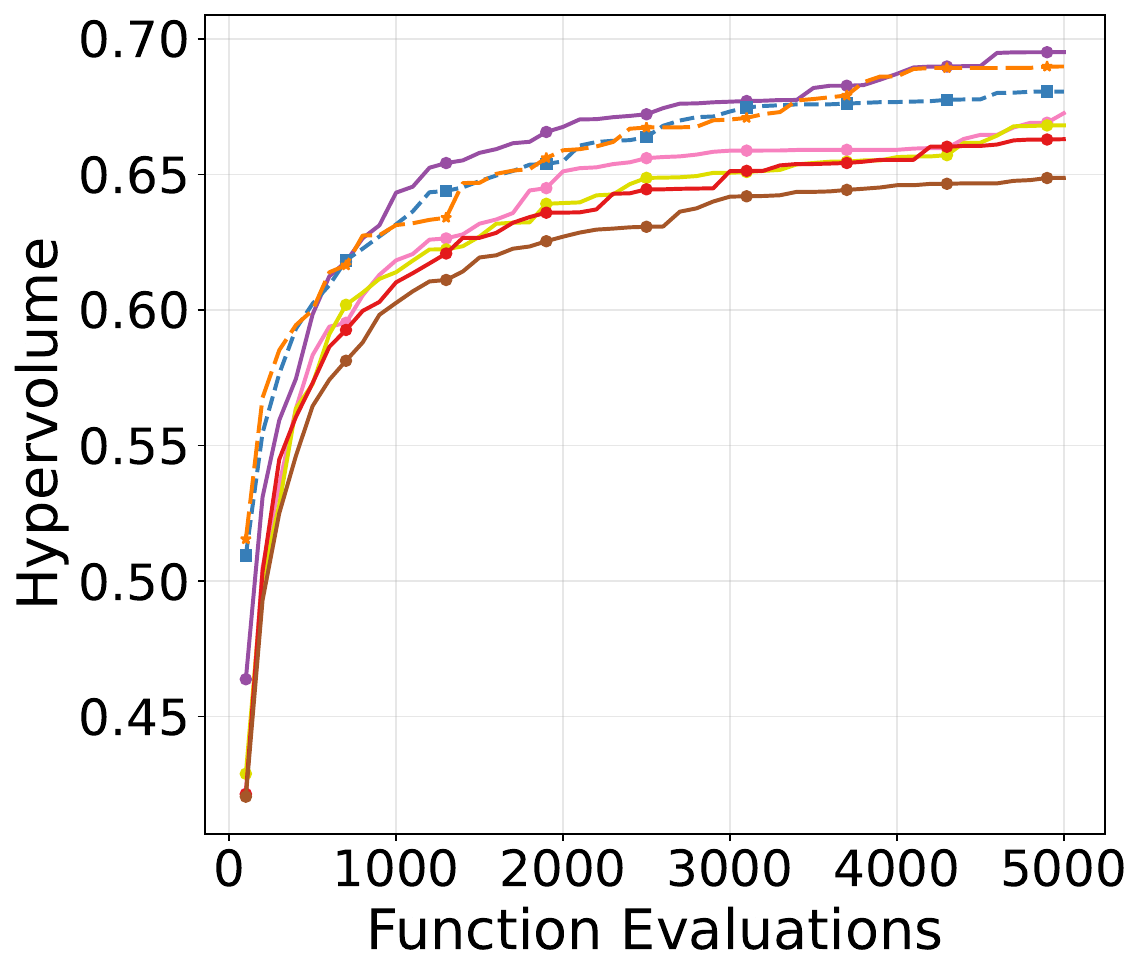}
        \caption{HV, lit. seeds}
        \label{fig:docking_lit_hv_curve}
    \end{subfigure}\hfill
    \begin{subfigure}[t]{0.24\linewidth}
        \centering
        \includegraphics[width=\linewidth]{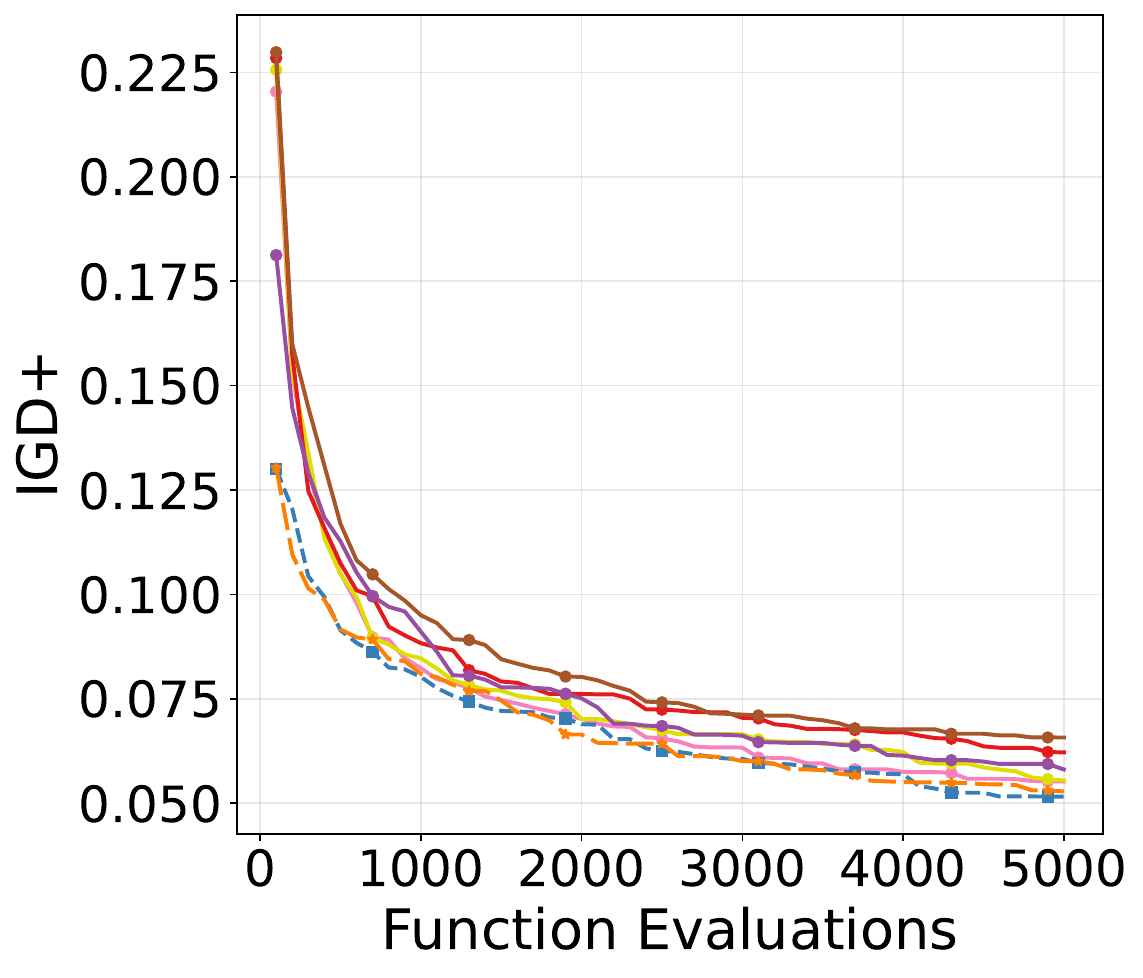}
        \caption{IGD$^+$, lit. seeds}
        \label{fig:docking_lit_igdplus_curve}
    \end{subfigure}
    \caption{Convergence on 6W63 docking: HV and IGD$^+$ against function evaluations under random initialization (a, b) and literature seeding (c, d). Each line is the median over runs for one configuration; colors match Fig.~\ref{fig:tradeoffs}, and run counts are in Table~\ref{tab:app_dock_medians}. Vertical scales differ between panels.}
    \label{fig:docking_curves}
\end{figure}